\documentclass{article} 
\usepackage{iclr2027_conference,times}
\usepackage{graphicx}
\usepackage{booktabs}
\usepackage{wrapfig}
\usepackage{amssymb}
\usepackage{titletoc}

\usepackage{amsmath,amsfonts,bm}

\def\eqref#1{equation~\ref{#1}}

\def\1{\bm{1}}

\DeclareMathAlphabet{\mathsfit}{\encodingdefault}{\sfdefault}{m}{sl}
\SetMathAlphabet{\mathsfit}{bold}{\encodingdefault}{\sfdefault}{bx}{n}

\usepackage{subcaption} 
\usepackage{hyperref}
\usepackage{url}
\usepackage{comment}

\usepackage{subcaption}
\usepackage[percent]{overpic}

\title{From Gist to Verbatim: The Controllable Path of One Image Through a Diffusion Prior}

\iclrfinalcopy

\author{%
  \parbox[t]{0.95\textwidth}{%
    \centering
    \vspace{1em}
    Luca Saglietti \quad Luca Giammanco \quad Marc Mézard\\[0.5em]
    Chiara Plizzari \quad Luca Biggio\\[0.8em]
    \textnormal{Universit\`a Bocconi}\\[0.3em]
    \textnormal{\texttt{firstname.lastname@unibocconi.it}}
  }
}
\newcommand{\chiara}[1]{\textcolor{magenta}{#1}}

\begin{document}

\maketitle
\lhead{}
\begin{abstract}

Modern generative models are typically studied for their ability to generalize, while memorization is treated as a failure mode. We invert this perspective and ask how a generative model can recall a single datum. We frame memorization as a control problem: a learned conditioning signal steers a frozen diffusion model toward a target image, and we follow the formation of this memory throughout training. Before converging to exact reproduction (\textit{verbatim}), memory passes through a regime of \emph{abstraction}: samples preserve the target's semantic content (\textit{gist}) while remaining diverse in appearance, leaving predictable detail to the generative prior. This semantic organization emerges despite the absence of labels, captions, or any explicit semantic objective. In the abstraction regime, the learned memory approaches ordinary text conditioning and supports editing, one-shot recognition, and retrieval more effectively than in the near-perfect reconstruction regime. These results suggest that memorization can serve as a probe of the relevance structure encoded by a generative model: what must be stored about a datum, what can be reconstructed from prior knowledge, and how learned memories interpolate between reconstructive recall and literal data storage.

\end{abstract}

\section{Introduction}

\begin{figure}[t]
\centering
\includegraphics[width=\textwidth]{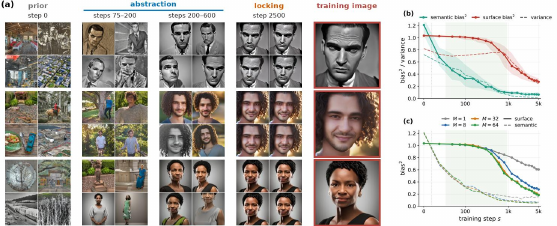}
  \vspace{-8pt}
\caption{\textbf{One image, two regimes: abstraction before locking.}
\textbf{(a)} Three memories ($M{=}8$ vectors each), each trained on a single image (rightmost, red frame) and used to condition a frozen diffusion model. Each $2{\times}2$ block shows four samples at different input noise realizations. At initialization, samples follow the prior (first column). During an extended \emph{abstraction} regime, the memory first generates other subjects sharing the target's style (second column), then the target's identity with varying pose, clothing and lighting (third column). By step 2500, \emph{locking} sets in and the training image is generated regardless of the input noise realization (fourth column).
\textbf{(b)} The same trajectory across $30$ single-image memories ($10$ animals, $10$ vehicles, $10$ faces; median$\pm$IQR). Solid: squared distance between the mean sample and the target (bias$^2$), measuring progress toward it. Dashed: dispersion of the samples around their mean (variance), measuring remaining variability. Semantic distance (DINOv2 embedding) collapses within ${\sim}10^2$ steps, while surface distance (VAE latent) collapses an order of magnitude later. Green band: the abstraction window, between the median semantic and surface half-decay steps (Appendix~\ref{app:exp_trajectory}).
\textbf{(c)} bias$^2$ by memory size (solid: surface, dashed: semantic). At fixed training budget a larger memory leaves the timescales unchanged but lowers the final surface residual. $M{=}1$ does not complete locking within the budget.
  \vspace{-10pt}
\label{fig:trajectory}}
\end{figure}

A scientific textbook condensing decades of research can occupy less disk space than a blurry vacation photograph. This fact may feel alien from a human perspective, because of the fundamental difference between data storage and memory. A disk stores any object literally, without a notion of relevance or an interpretative prior. Memory, instead, does not preserve an observation in its entirety: we make a choice about what to retain, and reconstruct the rest at recall from prior knowledge. \emph{How does the interaction between new information and prior shape a memory as it forms?}

This question sits at the intersection of two longstanding lines of work. The reconstructive view of memory distinguishes the stored trace from the prior knowledge used to rebuild an experience~\citep{bartlett1932remembering,schacter2007cognitive,spens2024generative}. Recent work in computational neuroscience interprets hippocampal traces as compressed cues for neocortical generative reconstruction and episodic--semantic memory as an adaptive compression process~\citep{spens2026hippocampo,nagy2025adaptive}. Coding and minimum-description-length perspectives, similarly, make description length relative to a given model: once the model is fixed, only the unexplained residual needs to be encoded~\citep{rissanen1978modeling,grunwald2019minimum}. Both views suggest that memory should be understood relative to what is already known. What remains unanswered, however, is how the information in a single datum is ordered by relevance: \emph{what is committed to memory first, and what can initially remain implicit in the prior.}

Modern generative models provide us with an experimental playground in which we can explore this question. Recent work on memorization in diffusion models has investigated when training examples are reproduced and how such replication can be detected or mitigated~\citep{carlini2023extracting,somepalli2023understanding}. In parallel, multiple personalization methods have been proposed to adapt pretrained diffusion models to new subjects or concepts so that they can be regenerated in novel contexts~\citep{gal2023image,ruiz2023dreambooth,kumari2023multi,voronov2023loss}. 
In this work, we ask a different question: if a strong generative prior is deliberately made to \textbf{memorize one image}, what does it learn along the way? \emph{How does memory formation unfold?} We freeze the prior, train only a conditioning signal from a single image, and follow the resulting memory throughout training.

The resulting trajectory is surprising yet familiar. Before approaching exact reconstruction, the memory passes through an \emph{abstraction regime}: generated samples incorporate the target's semantic content (\textit{gist}) while retaining diversity in pose, appearance, background, and other instance-specific details. Only after further training the memory enters a \emph{locking phase}, in which recall becomes \textit{verbatim}, i.e., the target image is fully reproduced.  No labels, captions, or semantic criteria are provided during memory formation. This dynamic is reminiscent of spectral bias, whereby neural networks preferentially learn some modes before others~\citep{rahaman2019spectral}. Here, however, the ordering is not simply one of spatial frequencies: it emerges from the interplay between a new datum and the structure already encoded by the pretrained generative prior. 
Memory recall becomes a probe of relevance: what the model stores, what it leaves to prior knowledge, and in what order. Our contributions are:

\begin{itemize}
    \item We identify and quantify an \emph{abstraction regime} in single-image memorization, Fig.~\ref{fig:trajectory}, in which semantic content is acquired substantially before instance-specific appearance. A similar phenomenology can also be traced in different settings and data modalities, Fig.~\ref{fig:abstraction_modalities}. 
    
    \item We formulate memorization as prior-mediated control and provide a minimal theoretical framework in which learning depends jointly on the residual information in the target and the susceptibility of the generative prior. We then retrace two qualitative predictions on the abstraction regime in our experimental setting, Fig.~\ref{fig:generalisation}.
    \item We show that abstraction is a distinct functional regime, Fig.~\ref{fig:overview}, by evaluating three capabilities of the learned memory: \emph{editability} (integrating new information at recall), \emph{one-shot recognition} (recognizing unseen realizations of the same subject), and \emph{text--memory retrieval} (accessing memories through textual cues). These capabilities emerge during abstraction and deteriorate as reconstruction approaches the verbatim target, revealing a trade-off between recall precision and memory reusability.

\end{itemize}

\section{Memorizing through a diffusion prior}
\label{sec:method}
\paragraph{Background on conditional diffusion models.}
Diffusion models generate samples from complex distributions by
learning to reverse a gradual noising process. Starting from a
data point $\bm x_0\sim p_0$, we define its noised version
$\bm x_t\sim p_t$ as
\begin{equation}\label{eq:forwardxt}
\bm x_t = \alpha_t\bm x_0+\sigma_t\bm\epsilon,
\qquad
\bm\epsilon\sim\mathcal N(\bm 0,\bm I),
\qquad t\in[0,T].
\end{equation}
In common variance-preserving formulations,
$\alpha_t\to0$ and $\sigma_t\to1$ as $t\to T$,
so that $p_T$ approaches a standard Gaussian.
This framework includes DDPMs~\citep{ho2020denoising}
and continuous-time diffusion models~\citep{song2021scorebased}.

Since $p_T$ is easy to sample from, generation can be
performed by approximately reversing the noising process.
The reverse dynamics are determined by the score function
$\bm s(\bm x,t)=\nabla_{\bm x}\log p_t(\bm x)$.
As discussed in Appendix~\ref{app:derivation}, this
score can be learned through denoising score matching,
by optimizing a neural network $\bm s^\theta(\bm x,t)$
with the objective
\begin{equation}\label{eq:scorematchingL}
\mathcal L(\theta)
=
\mathbb E_{t,\bm x_0,\bm\epsilon}
\left[
\lambda(t)
\left\|
\bm s^\theta(\bm x_t,t)
+\frac{\bm\epsilon}{\sigma_t}
\right\|^2
\right],
\end{equation}
where $\lambda(t)$ allows different noise levels to
be weighted differently. The optimal predictor recovers
the marginal score $\bm s(\bm x,t)$.

In conditional diffusion models, generation is additionally
guided by a variable $\bm C$, such as a text caption.
Training on pairs of data and conditioning signals allows
the network to approximate the conditional score
$\bm s_{\bm C}(\bm x,t)
=\nabla_{\bm x}\log p_t(\bm x\mid\bm C)$,
and thus to generate samples from
$p_0(\bm x\mid\bm C)$.
\paragraph{Memory as a conditioning signal.} Consider a pretrained conditional diffusion model with data distribution \(p_0\) and score \(\bm s_0(\bm x,t)\). We study how such a model can be made to recall a single target image \(\bm x^\star\). Importantly, we aim to achieve this goal not by fine-tuning the parameters of the generative model, which are kept frozen at their pretraining values, but by learning a compact conditioning signal $\bm C$. In the following, we focus on the training trajectory of $\bm C$, which we regard as the \emph{memorization} process, and we refer to $\bm C$ as the \emph{memory}. During generation, the memory acts as a \emph{control}: it steers the reverse diffusion toward \(\bm x^\star\), while relying on the structure encoded by the pretrained prior.
\vspace{-0.2cm}
\paragraph{Residual to a prior.} The target has negative log-likelihood \(-\log p_0(\bm x^\star)\) under the pretrained generative model, which can be interpreted as its ideal codelength under the prior \citep{rissanen1978modeling,grunwald2019minimum}.\footnote{Strictly, this coding interpretation assumes a discretized or finite-precision representation.} In this view, the memory \(\bm C\) need not encode the datum from scratch: it should only supply what is not already explained by the model. Learning can therefore be viewed as progressively reducing a residual relative to the prior. The central question of this work is how this residual is resolved along the training trajectory: which properties of the datum are \emph{relevant} under the prior and stored first, and which are initially left for the model to guess.
\vspace{-0.2cm}

\paragraph{Training objective.} We learn the control signal $\bm C$ using the diffusion model's denoising objective on a single target image $\bm x^\star$, keeping all pretrained parameters frozen. Under the noising in \eqref{eq:forwardxt}, the noised target's exact score is
\begin{equation}
\bm s^\star(\bm x_t^\star,t)
=
-\sigma_t^{-2}(\bm x_t^\star-\alpha_t\bm x^\star).
\label{eq:scorestar}
\end{equation}
Let $\bm s_{\bm C}(\bm x,t)$ denote the frozen model's score under memory conditioning $\bm C$. We model \emph{memorization} by minimizing, with respect to $\bm C$ only,
\begin{equation}
\mathcal L(\bm C)
=
\mathbb E_{t,\bm\epsilon}
\left[
\lVert
\bm s_{\bm C}(\bm x_t^\star,t)
-
\bm s^\star(\bm x_t^\star,t)
\rVert^2
\right].
\label{eq:memory_objective}
\end{equation}
\vspace{-18pt}

\paragraph{Instantiation.} We instantiate this construction primarily in Stable Diffusion~2.1 (SD2.1) ~\citep{rombach2022high}. This is a latent diffusion model: a pretrained VAE maps an image to a latent representation, and a UNet learns to denoise noisy versions of that latent. Thus, we apply the denoising objective (\eqref{eq:memory_objective}) to the target's latent \(\bm z^\star=\mathrm{VAE}(\bm x^\star)\), using the model's native \(v\)-prediction parameterization and uniform time sampling. 

SD2.1 is text-conditioned. A text encoder maps a caption to a sequence of token embeddings, which condition the UNet through cross-attention: image features provide queries, while the text embeddings provide keys and values. When no caption is supplied, the model instead receives the embedding of the empty caption $\bm c_{\varnothing}$. We take this empty-caption model as our generative prior $p_0$.

We represent the memory by \(M\) learnable vectors \(\bm C=\{\bm c_1,\ldots,\bm c_M\}\), each of dimension \(D=1024\). These vectors enter a second cross-attention pathway parallel to the text pathway, using the same frozen projection parameters. Thus, at each UNet cross-attention layer,
\begin{equation}
\mathrm{Attn}_{\mathrm{tot}}(\bm h)
=
\mathrm{Attn}_{\mathrm{text}}(\bm h,\bm c_{\varnothing})
+
\mathrm{Attn}_{\mathrm{mem}}(\bm h,\bm C).
\end{equation}
Only \(\bm C\) is optimized; all parameters of the VAE, UNet, text encoder, and attention projections remain frozen. We initialize \(\bm C\) with small random values so that the memory contribution is initially negligible and the model starts close to the prior \(p_0\). Notice that our learned conditioning vector is fed to the cross-attention mechanism \emph{directly}, unlike the user prompt that has to pass through a text encoder. This allows our setting to be as agnostic as possible to the fact that SD2.1 is conditioned through text, and to be generalizable to any model that can be conditioned through attention.

\vspace{-0.2cm}
\paragraph{Model-generated vs. external targets.} Unless stated otherwise, our main experiments use targets generated by SD2.1 from a curated set of captions. We make this choice for two reasons. First, targets within the model's generative repertoire should have smaller initial residuals and permit lower reconstruction bias in the asymptotic training regime of \(\bm C\). Second, each target comes with a known semantic reference, the caption it was generated from, which we use in subsequent analyses. Crucially, this caption is not provided during memorization: \(\bm C\) is learned from the target image alone. In Appendix~\ref{app:add_ood}, we repeat our experiments on external data.

\section{From gist to verbatim}
\label{sec:gist_to_verbatim}

We first ask what a memory acquires, and in which order. As displayed in Fig.~\ref{fig:trajectory} (a), following a single-image memory throughout training (minimizing \eqref{eq:memory_objective}) reveals a consistent trajectory. 

\paragraph{Abstraction then Locking.} At initialization, samples reproduce the pretrained prior. After relatively few updates, generations become semantically aligned with the target while remaining diverse in pose, appearance, background, and other instance-specific details. With further training, and a growing norm of the memory vectors (see Appendix~\ref{app:add_norm}), this variability progressively disappears until different noise realizations converge toward the same image. We call the intermediate phase the \emph{abstraction regime}: the memory has acquired the target's semantic content while its particular appearance remains substantially unconstrained. This semantic organization is not imposed by the memory-training objective, which provides neither labels nor semantic supervision. Abstraction is therefore not the emergence of semantics from scratch, but a resonance between datum features and semantic structure already represented by the prior. We call the later training stage the \emph{locking phase}, in which realization-specific spatial and surface details are progressively fixed to the target.

\vspace{-0.2cm}
\paragraph{Semantics vs Surface.}
We quantify this transition in Fig.~\ref{fig:trajectory} (b). At each checkpoint, we
generate multiple samples from independent initial noises and compare them with the target in
DINOv2~\citep{oquab2024dinov2learningrobustvisual} (semantic) and VAE (surface) representation spaces. We measure two quantities: the
squared distance between the mean sample and the target representations (\emph{bias}$^2$), and
the dispersion of individual samples around their mean (\emph{variance}). Both are normalized
by the squared norm of the target representation; exact definitions are given in
Appendix~\ref{app:exp_trajectory}.

The two types of information are acquired on markedly different timescales. The median ratio of surface to semantic half-decay steps, defined as the step at which each bias$^2$ curve has completed half of its observed decrease, is $21.4$ (Fig.~\ref{fig:trajectory}b). Semantic bias drops rapidly and is largely removed during the early phase of training, whereas surface bias remains high over the same interval and decreases only later, in the locking phase. Importantly, while semantic bias has largely converged by the end of training, surface bias often has not. If the latter continues to decrease, the finite training endpoint underestimates its half-decay step. Since our interest is in establishing the ordering rather than the precise timescale ratio, this makes our conclusion robust to the finite training horizon.
The semantic-first ordering is robust across alternative semantic and surface representations, normalization choices, and independent training initializations (Appendices~\ref{app:moremetrics} and~\ref{app:restarts}).
In Appendix~\ref{app:add_ood}, we also show that the same phenomenology extends to external photographs. For strongly out-of-distribution targets, however, the abstraction regime can weaken or disappear altogether, depending on the properties represented by the prior (Appendix~\ref{app:add_ood_strong}).

\vspace{-0.2cm}
\paragraph{Impact of memory size.} Memory size, \(M\), primarily affects the locking phase. As shown in Fig.~\ref{fig:trajectory}(c), increasing the number of learned vectors lowers the final surface error, but the early semantic timescale is largely unaffected. In particular, a single-vector memory (\(M=1\)) acquires semantic content on approximately the same schedule as larger memories but does not approach exact reconstruction. Abstraction is therefore not simply a consequence of insufficient representational capacity: even memories that ultimately reproduce the target pass through a phase in which semantic content has been acquired while finer details remain unresolved. The role of memory norm, including rescaling and fixed-norm controls, is examined in Appendix~\ref{app:add_norm}.

\section{A framework for abstraction as prior-mediated control}
\label{sec:framework}

To organize these observations, we introduce a minimal mathematical model that captures conditions for an abstraction regime to emerge. Let \(\bm\phi:\mathbb{R}^d\rightarrow\{0,1\}^F\) be a collection of measurable binary features capturing properties represented by the generative model. These features may range from high-level semantic attributes (e.g., ``a face'') to spatially localized geometric details (e.g., ``a particular pose at a particular image location''). 
We model the effect of a learned memory as an exponential tilt of the pretrained generative distribution,
\begin{equation}
p_{\bm u}(\bm x) = \frac{
    e^{\bm u^\top\bm\phi(\bm x)}
}{ Z(\bm u) } p_0(\bm x), \qquad
Z(\bm u) = \mathbb{E}_{p_0} \left[
    e^{\bm u^\top\bm\phi(\bm X)} \right].
\label{eq:feature_tilt}
\end{equation}
Here, \(\bm u\in\mathbb R^F\) is the theoretical counterpart of the learned memory \(\bm C\). While \(\bm C\) is represented by learned conditioning vectors in SD2.1, \(\bm u\) directly weights the abstract feature coordinates \(\bm\phi\), determining which properties of the target are emphasized. The tilted score decomposes as
\(\bm s_{\bm u}(\bm x,t)=\bm s_0(\bm x,t)+\delta\bm s_{\bm u}(\bm x,t)\), with
\begin{equation}\label{eq:controlformula}
\delta\bm s_{\bm u}(\bm x,t)
=
\nabla_{\bm x}
\log
\mathbb{E}_0
\!\left[
e^{\bm u^\top\bm\phi(\bm X_0)}
\,\middle|\,
\bm X_t=\bm x
\right],
\end{equation}
where \(\mathbb{E}_0[\cdot\mid\bm X_t=\bm x]\) denotes expectation under the posterior induced by \(p_0\) and the forward noising process. 
The correction \(\delta\bm s_{\bm u}\) is a state- and time-dependent control field: at each noise level, it evaluates under the pretrained model's own posterior how far the demanded properties are from being realized, and pushes the dynamics accordingly. 

To study how the memory is acquired, define the current feature residual
\begin{equation}
\bm r_{\bm\phi}(\bm u)
=
\bm\phi(\bm x^\star)
-
\mathbb{E}_{p_{\bm u}}
\left[
\bm\phi(\bm X)
\right].
\label{eq:feature_residual}
\end{equation}
For the variance-preserving diffusion process and an appropriate diffusion-time weighting (see Appendix~\ref{app:derivation}), the denoising objective becomes equivalent to minimizing the target negative log-likelihood. Although this weighting differs from that used in SD2.1, it allows us to obtain simple gradient-flow dynamics for the control \(\bm u\):
\begin{equation}\label{eq:gradientflow}
\frac{d\bm u}{d\tau}
=
-\eta\nabla_{\bm u}\mathcal{L}
=
\eta\,\bm r_{\bm\phi}(\bm u),
\qquad
\frac{d\bm r_{\bm\phi}}{d\tau}
=
-\eta\bm\Sigma_{\bm u}\bm r_{\bm\phi},
\end{equation}
where the feature covariance \(\bm\Sigma_{\bm u}=\operatorname{Cov}_{p_{\bm u}}[\bm\phi(\bm X_0)]\) under the tilted model is a linear \emph{susceptibility}, since \(\partial_{\bm u}\mathbb E_{p_{\bm u}}[\bm\phi]=\bm\Sigma_{\bm u}\): feature-space directions with larger eigenvalues respond more strongly to infinitesimal changes in the tilt parameters \(\bm u\). The memorization dynamics in~\eqref{eq:gradientflow} are governed by two ingredients: the residual \(\bm r_{\bm\phi}\), specifying which properties of the target are not yet reproduced, and the \emph{susceptibility} \(\bm\Sigma_{\bm u}\), determining which feature-space directions can change most readily.

Considering a simple binary feature helps the intuition. If a feature is present with probability \(p\) under the current tilted model, its susceptibility is \(\operatorname{Var}(\phi)=p(1-p)\). A highly specific event (``a pose at a location'') has \(p\ll1\), and hence low susceptibility. A broader property (``a face'') can have substantially larger susceptibility and therefore be acquired more readily. In the multivariate setting, correlations add another source of susceptibility: semantic properties often participate in coherent combinations, while highly localized details are typically less strongly coupled to other features. This is consistent with Fig.~\ref{fig:trajectory}, where identity and style emerge while pose and clothing still vary.

The framework is deliberately minimal: its purpose is to identify the \emph{ingredients that can give rise to an abstraction regime}, rather than to provide an exact description of SD2.1. In particular, we do not claim that exponential tilting precisely reproduces the effect of cross-attention conditioning. Therefore, showing that high-susceptibility semantic modes can lead to abstraction does not establish that this mechanism is responsible for the behavior observed in our experiments. Direct verification of this hypothesis is challenging given the complexity of SD2.1. Nevertheless, the framework yields two qualitative predictions that we can test experimentally.
(i) Early in memorization, the tilt is weak, so the susceptibility structure remains close to that of the pretrained prior. Through pretraining, text conditioning aligns with high-susceptibility semantic directions encoded in the prior. Yet captions do not fully constrain generated images: different initial noises produce different realizations consistent with the same text. If an abstract memory acts as a similarly weak semantic constraint, it should preserve some of this noise-dependent prior variability. (ii) As the fast semantic residuals are removed, learning shifts toward slower, low-susceptibility features. Exact reconstruction must capture these finer constraints, including those that specify and control the spatial realization of the target. In our implementation, this should appear as increasingly position-dependent addressing.

\section{Analysis of the abstraction regime}
\label{sec:analysis}

Section~\ref{sec:framework} suggests two signatures of the abstraction and locking phases. 
First, an abstract memory should preserve variation inherited from the prior, yielding caption-like behavior during this phase. Second, as the memory fixes spatial details during locking, its influence should become more position-specific. In the following, we test these predictions.


\subsection{Generalization through the prior}

To test whether memory preserves variation inherited from the prior, we generate a collection of images from a fixed caption using different initial noise realizations, then train a single memory jointly on $N$ of these images. For $N>1$, this becomes a form of prompt inversion. We evaluate generalization on held-out initial noises, comparing, for each noise, the images generated under the original caption and the learned memory. The same test remains meaningful at $N=1$: systematic correspondence across held-out noises indicates that the memory recovers, through the prior, variation absent from its single training image.

\vspace{-0.2cm}
\paragraph{Paired advantage.}
We measure this correspondence using DINOv2 distance between memory- and
caption-conditioned samples. For each held-out initial noise, we compare the distance between
the two samples generated from the same noise with the average distance obtained by pairing
the memory sample with caption-conditioned samples generated from different noises. We call
the unpaired-minus-paired distance the \emph{paired advantage}, normalized by the mean
pairwise distance. Further details on the metric formulation are provided in Appendix~\ref{app:exp_generalisation}, Equation \ref{eq:advantage}.

At initialization, memory conditioning is negligible and the prior alone differs substantially from caption conditioning, so sharing the same initial noise provides little correspondence and the advantage is near zero. During abstraction, a positive advantage indicates that memory and caption conditioning steer the prior similarly enough for the same initial noise to produce corresponding realizations. As exact recall is approached, memory conditioning suppresses this noise dependence, causing the advantage to return toward zero.
Details in Appendix~\ref{app:exp_generalisation}.
\vspace{-0.2cm}

\begin{figure}[t]
  \centering
  \includegraphics[width=0.9\textwidth]{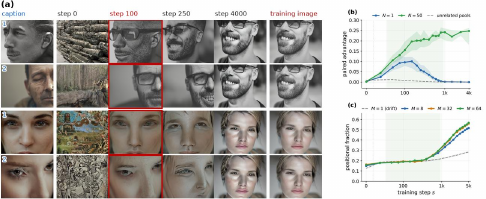}
\caption{\textbf{Generalization through the prior.}
\textbf{(a)} Each row corresponds to a caption--noise pair. For each of two captions, we show two input-noise realizations (1-2). Columns: the caption-conditioned sample; memory-conditioned samples ($N{=}1$) at initialization, near the peak of the paired advantage (step 100, red frame), later in training (step 250) and after locking (step 4000); the training image. The input noise is held fixed along each row. For each caption, the two realizations shown are those with the largest paired advantage at the peak.
\textbf{(b)} Paired advantage over training for twelve captions describing faces, one memory per caption at $N{=}1$ (blue) and $N{=}50$ (green); median, bands: IQR. Dashed grey: the same samples scored against pools from unrelated captions at the same input noise.
\textbf{(c)} Positional fraction $f$ over training, mean over $30$ memories per memory size ($10$ each of animals, vehicles and faces; bands: s.e.m.). Dashed: the $M{=}1$ control, which cannot directly learn position-dependent addressing.
\vspace{-10pt}
\label{fig:generalisation}}
\end{figure}

\paragraph{Generalization from a single image.}
Figure~\ref{fig:generalisation} (a-b) shows two examples and the aggregate result over twelve
captions. At initialization, memory samples are unrelated to the caption-conditioned
references. Near the paired advantage peak, samples tend to follow properties of their own paired
reference, including pose, framing, and color, while remaining distinct from the
training image. Later in training, some of this variation remains, but the samples
increasingly acquire the target's identity, palette, and background. This intermediate form
of \emph{biased generalization}~\citep{garnier2026biased,rodriguez2026diffusion} is consistent with instance-specific constraints strengthening after the semantic content has been acquired. With further training, dependence on the initial noise disappears as the outputs converge toward the
memorized image, and the paired advantage returns to zero.

At the peak, the paired caption reference is the nearest neighbour for about $20\%$ of memory-conditioned samples (chance level is $2\%$ in this experiment). The effect is therefore substantial but not uniform across noise realizations, showing a restriction of the invariance supported by the caption-conditioned prior. The same samples show no advantage when compared with images from
unrelated captions, while memories trained on $N=50$ images from the same caption reach roughly twice the peak advantage and preserve the correspondence later in training.

\subsection{From content dependence to positional specificity}

The second prediction concerns how the memory acts within the network. Memory tokens influence generation through cross-attention, where they are projected into keys and values. Query--key interactions determine \emph{where} each token contributes, while values determine \emph{what} information it supplies. We therefore examine how memory's addressing changes during training.

\vspace{-5pt}
\paragraph{Content and position in the attention logits.}
We want to distinguish whether the memory's addressing depends on the image being processed
or instead becomes tied to fixed spatial locations. We examine the pre-softmax
interaction between image queries and memory keys. We use the cosine similarity between image queries and memory keys, which factors out the large change in memory norm during training and remains
informative even for $M=1$, where the softmax attention weight is identically one.

Using a fixed set of held-out images, we decompose the variability of these scores by the law
of total variance. The first component, $V_{\mathrm{cont}}$, measures how much the score
varies across images at fixed spatial positions. The second, $V_{\mathrm{pos}}$, measures the
variation across positions that remains after averaging over images. We report the
\emph{positional fraction}, $f$, the fraction of the total variance accounted for by
$V_{\mathrm{pos}}$. Small values therefore indicate predominantly content-dependent
addressing, whereas large values indicate increasing positional specificity. Further details
are given in Appendix~\ref{app:exp_positional}.

\paragraph{Addressing becomes position-specific during locking.}
Figure~\ref{fig:generalisation} (c) shows a clear transition. Throughout the abstraction regime, the
positional fraction remains low and approximately constant. As locking sets in, it rises
sharply for multi-token memories, indicating that a spatial pattern shared across images
accounts for an increasing fraction of the addressing variability.

The $M=1$ case provides a useful control. With a single memory token, the softmax attention
weight is always one, so positional attention cannot be learned directly. The shared memory vector is still updated through the value pathway, which can indirectly change the query--key logits and produce a gradual positional drift. This can support good reconstruction through the frozen prior, but produces only the gradual change seen for \(M=1\). The sharp increase in positional specificity appears only for \(M>1\) as memorization approaches the verbatim regime.

\section{The fidelity--functionality trade-off} \label{sec:func-fidel}

Memory theories have long distinguished gist-like from verbatim representations, and behavioral
work suggests that generalization or accessibility can come at the cost of precision
\citep{brainerd2002fuzzy,berens2020dissociating}. In diffusion personalization, observations of early saturation in customization quality have motivated efficient early-stopping strategies~\citep{voronov2023loss}. We now ask whether a fidelity--functionality trade-off emerges
spontaneously during the formation of our memory. In particular, does increasing
fidelity to the stored instance necessarily improve what the memory can do?

\begin{figure*}[t]
  \centering
  \begin{subfigure}[t]{0.32\textwidth}
    \centering
    \begin{overpic}[width=\linewidth]
    {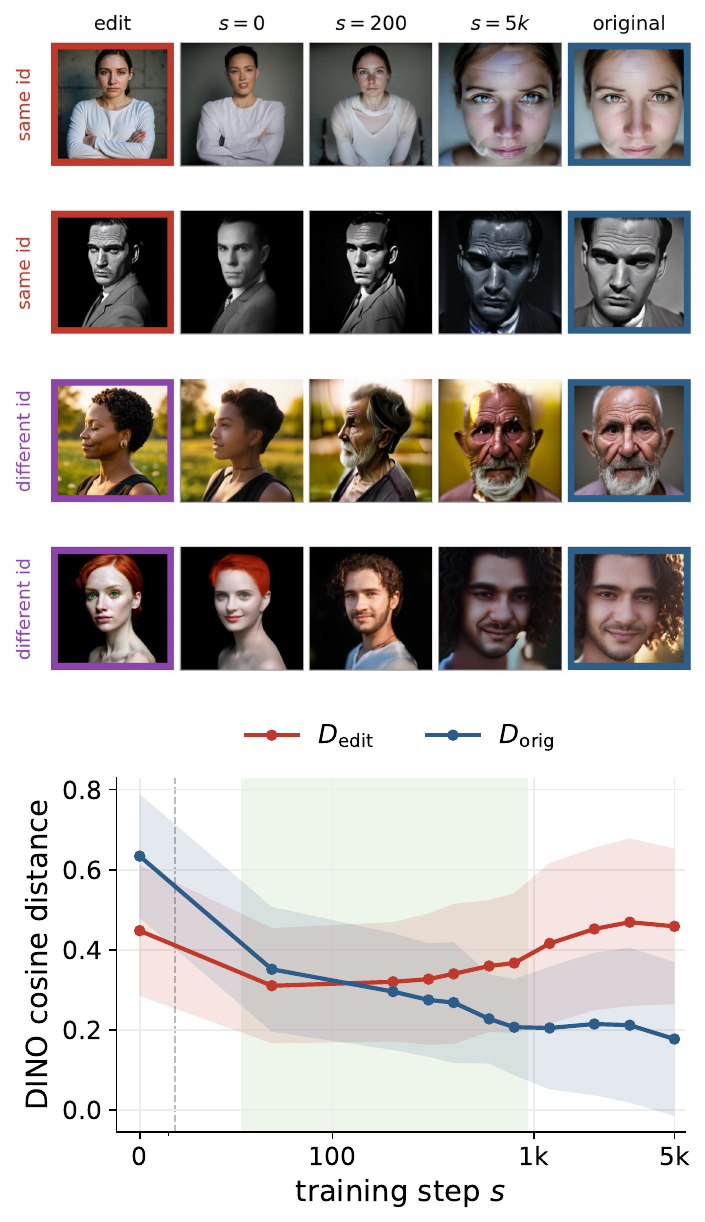}
      \put(2,03){\makebox(0,0)[tl]{\textbf{(a)}}}
    \end{overpic}
    \phantomsubcaption
    \label{fig:editability}
  \end{subfigure}\hfill
  \begin{subfigure}[t]{0.32\textwidth}
    \centering
    \begin{overpic}[width=\linewidth]
    {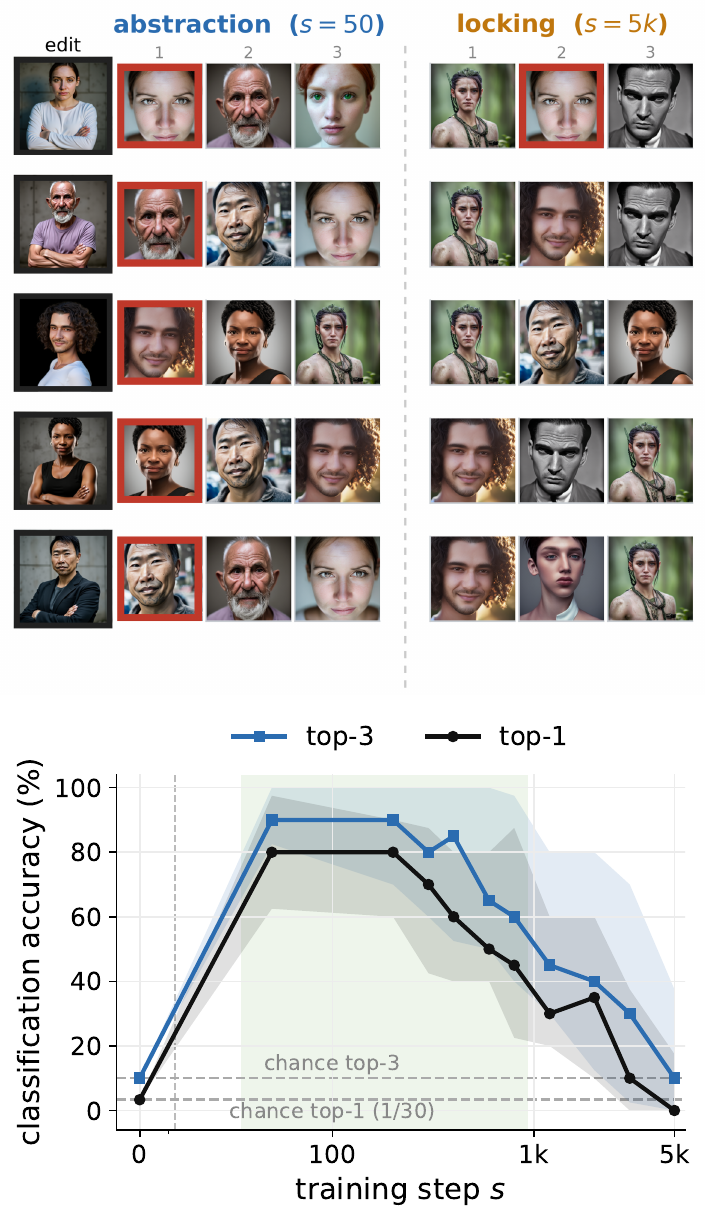}
      \put(2,03){\makebox(0,0)[tl]{\textbf{(b)}}}
    \end{overpic}
    \phantomsubcaption
    \label{fig:zeroshot}
  \end{subfigure}\hfill
  \begin{subfigure}[t]{0.32\textwidth}
    \centering
    \begin{overpic}[width=\linewidth]
    {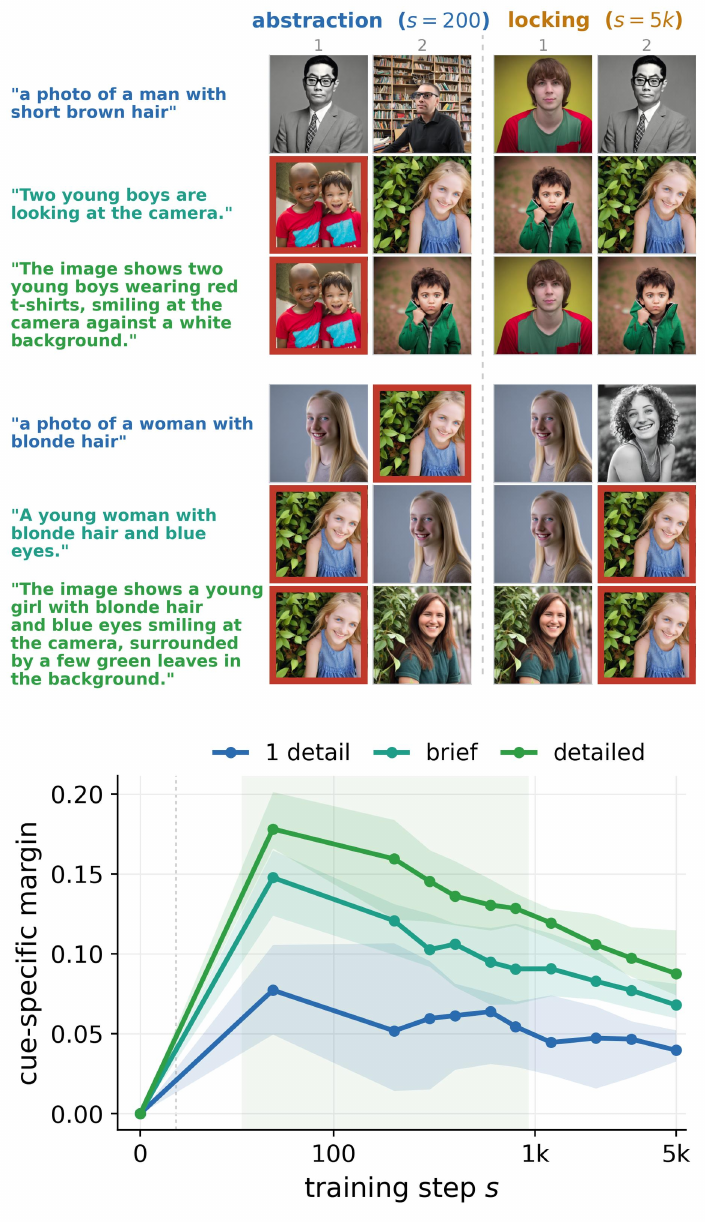}
      \put(2,03){\makebox(0,0)[tl]{\textbf{(c)}}}
    \end{overpic}
    \phantomsubcaption
    \label{fig:retrieval}
  \end{subfigure}
  \vspace{-15pt}
\caption{\textbf{Fidelity--functionality trade-off.}
\textbf{(a) Editability.} Edited images are partially noised and denoised under the target's memory.
Top: two same-identity edits and two cross-identity inputs. Columns show the edit, outputs at three checkpoints, and the target.
Outputs retain edits during abstraction and revert toward the target during locking.
Bottom: DINOv2 distance to the edit ($D_{\mathrm{edit}}$) increases as distance to the target ($D_{\mathrm{orig}}$) decreases, measured over 20 same-identity target--edit pairs with two samples per pair.
\textbf{(b) One-shot recognition.} Each of 300 held-out images (10 per identity) is classified among the 30 identity memories, assigned to whichever yields the lowest denoising error over 20 shared timestep--noise pairs (chance: 1/30).
Top: the three best-scoring identities (of 30) for five queries (edits) during abstraction and locking.
Bottom: top-1 and top-3 accuracy (median and IQR over identities), over 10 queries per identity.
\textbf{(c) Text--memory retrieval.} Each of 15 memories is trained without text cues on a single image generated from a detailed caption.
Cues at three specificity levels rank memories by cosine similarity between text-encoder EOS embeddings and mean-pooled memory vectors.
Top: the two highest-ranked memories for two identities during abstraction and locking, represented by their training images.
Bottom: similarity to the correct memory minus mean similarity to the other memories.
In (b) and (c), red frames indicate correct matches.
Curves: medians, bands: IQR across target--edit pairs in (a) and identities in (b--c).}\label{fig:overview}
  \vspace{-10pt}
\end{figure*}

We probe three complementary forms of functionality. 
\textit{Editability} asks whether the memory
can be integrated with new information supplied at recall. \textit{One-shot recognition} asks whether it retains useful invariance and can provide a good explanation across unseen realizations of the same subject. \textit{Text--memory retrieval}
asks whether it can be accessed through a textual cue. Across all three probes,
functionality emerges during abstraction and subsequently deteriorates as reconstruction approaches
the verbatim target (Fig.~\ref{fig:overview}), even as recall fidelity continues to improve.

\vspace{-0.2cm}

\paragraph{Editability.}
We follow a protocol analogous to diffusion-based editing, which starts from a partially noised input and lets the prior complete the denoising process~\citep{meng2021sdedit}. For each target $x_{\mathrm{orig}}$ and edited counterpart $x_{\mathrm{edit}}$
(generated with FLUX~\citep{batifol2025flux1kontext}; see Appendix~\ref{app:exp_edit}), we partially noise the edit and complete denoising
under the checkpointed memory, following a U-turn protocol. We then measure the resulting sample's DINOv2 cosine distance from the edit,
$D_{\mathrm{edit}}$, and from the stored image, $D_{\mathrm{orig}}$. During abstraction, changes in pose, expression, accessories, and
scene remain visible in the output. As memorization progresses, $D_{\mathrm{edit}}$ increases while
$D_{\mathrm{orig}}$ decreases: the same memory that reconstructs its target more faithfully becomes
progressively less compatible with the external edit (Fig.~\ref{fig:editability}). The same transition
occurs when the new information to be integrated is an image of a different person: an abstract memory preserves substantial information
from the input, whereas a verbatim memory increasingly overrides it with its stored subject.
Sampling, noising, and averaging details are given in Appendix~\ref{app:exp_edit}.
\vspace{-0.2cm}
\paragraph{One-shot recognition.}
Conditional denoising error can itself be used as a generative classifier, both in zero-shot 
\citep{li2023your} and in few-shot
\citep{chattopadhyay2024towards} settings. Here, 
we ask whether our trained memories transiently acquire
enough invariance to recognize unseen realizations of their corresponding subject. We train one memory for each of thirty distinct identities (human faces), and construct a held-out
query set showing the same identities in different poses and scenes. At each checkpoint, we evaluate the denoising loss in~\eqref{eq:memory_objective} for each query under
every candidate memory, and assign the query to the identity whose memory gives the lowest loss. Recognition accuracy rises rapidly during abstraction: top-3 accuracy reaches $0.9$ and top-1 accuracy $0.8$,
both following the same non-monotonic trajectory, before deteriorating towards chance as
locking sets in (Fig.~\ref{fig:zeroshot}). Thus, while reconstruction of the training image
continues to improve, the memory progressively loses the invariance required to recognize the same
subject under new appearances.
\vspace{-0.2cm}

\paragraph{Text--memory retrieval.}
While text never enters memory formation, we can still ask whether a memory trained from an image alone becomes accessible through semantic cues, and how this accessibility changes as recall becomes more precise. For each identity, we generate a single training image from a detailed caption and train a corresponding memory. At evaluation time, we query the memory with three textual descriptions
of increasing specificity: a category label plus one distinguishing attribute, a brief description, and a detailed description of the image (see Appendix~\ref{app:exp_retrieval}). At each checkpoint, we mean-pool the $M$ conditioning vectors and compare the resulting representation with the frozen caption embeddings by cosine similarity. For each cue
level, retrieval is given by the similarity between each description and the corresponding identity, measured relative to the other identities. Alignment with the semantic cues emerges rapidly and peaks in the abstraction regime, despite text never entering the memory-training objective. More detailed cues retrieve the memory
more strongly, while all levels lose alignment as the memory becomes increasingly specialized to the particular training realization (Fig.~\ref{fig:retrieval}).

These experiments reveal a fidelity--functionality trade-off: as recall becomes more precise, memory becomes less editable, less invariant across realizations, and less accessible through semantic cues. Additional instance-specific information thus changes memory functionality.

\section{Extensions to Other Models and Modalities}
\vspace{-5pt}

\begin{figure}[t]
    \centering
    \begin{subfigure}[b]{0.49\linewidth}
        \centering
        \begin{overpic}[width=\linewidth]{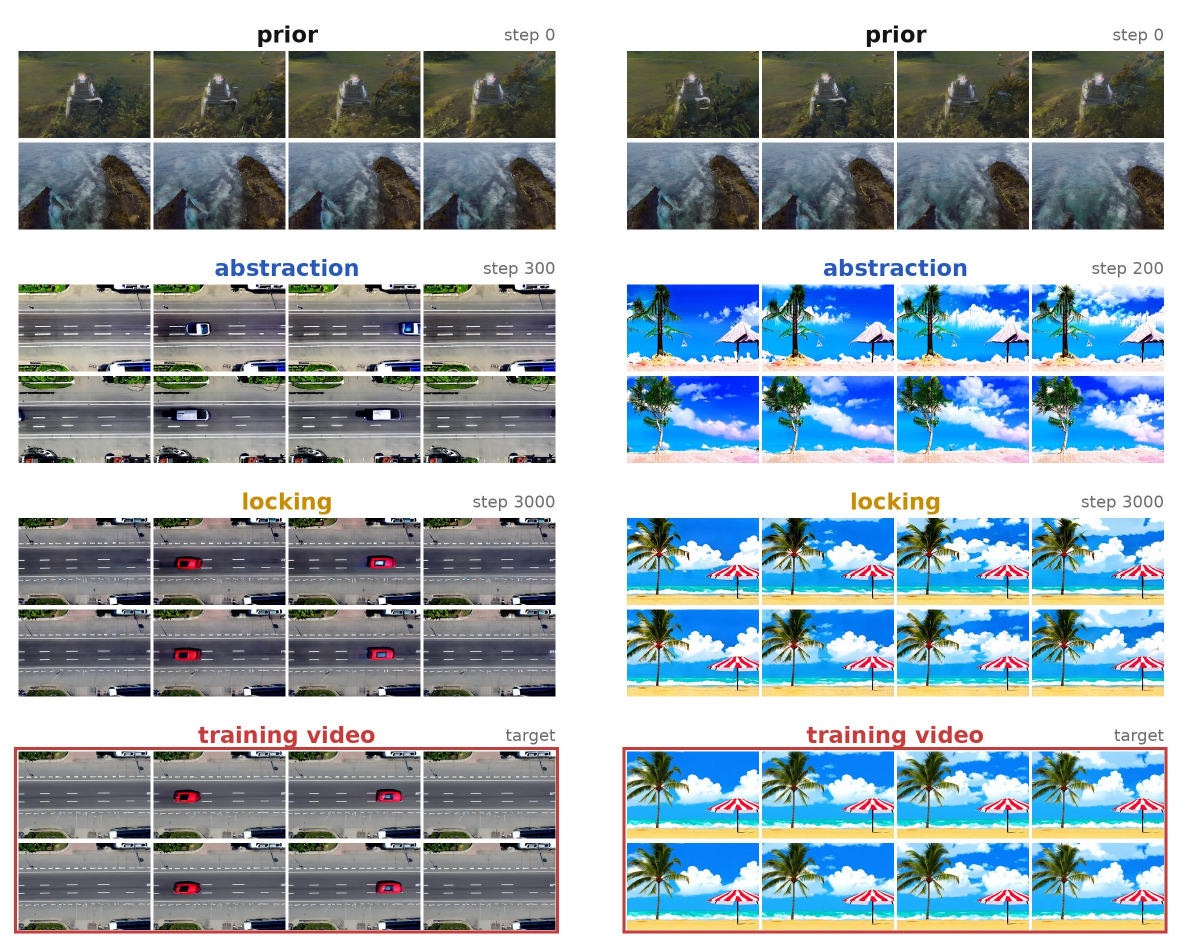}
            \put(1,82){\makebox(0,0)[tl]{\textbf{(a)}}}
        \end{overpic}
        \phantomsubcaption
        \label{fig:video}
    \end{subfigure}
    \hfill
    \begin{subfigure}[b]{0.49\linewidth}
        \centering
        \begin{overpic}[width=\linewidth]{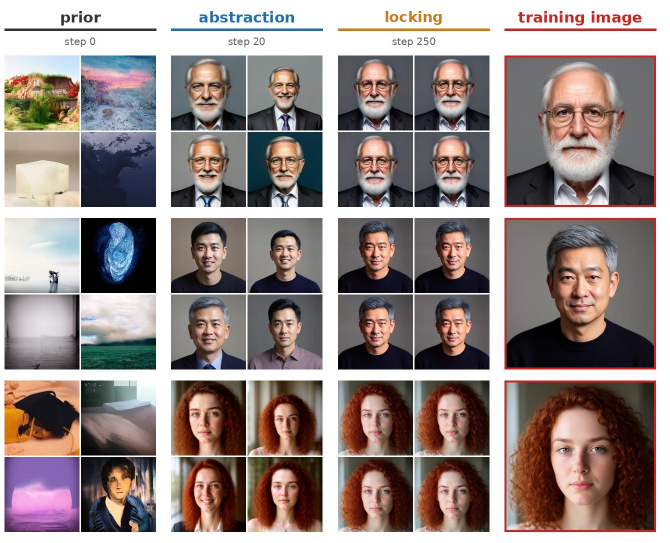}
            \put(1,82){\makebox(0,0)[tl]{\textbf{(b)}}}
        \end{overpic}
        \phantomsubcaption
        \label{fig:faces}
    \end{subfigure}
        \vspace{-15pt}
\caption{\textbf{Abstraction across models and modalities.}
\textbf{(a)}~Video memory on CogVideoX-2b, each block shows two seeds as filmstrips. During abstraction, characteristic motion, e.g., motion of the car and moving direction of the clouds, stabilize while frame content and layout vary. Finally, every seed reproduces the same target.
\textbf{(b)}~Image memories on FLUX.2 for three identities (rows), each sampled with four seeds. Semantic content emerges well before verbatim appearance.}
\label{fig:abstraction_modalities}
    \vspace{-10pt}
\end{figure}

We now test whether the gist-to-verbatim ordering observed in SD~2.1 extends across architectures, generative objectives, memory parameterizations, and modalities. We repeat single-datum memorization with CogVideoX-2b~\citep{yang2024cogvideox} and the rectified-flow image model FLUX.2~\citep{flux2024}, adapting memory injection to each model's native attention architecture.
\vspace{-0.2cm}
\paragraph{Abstraction across architectures and modalities.}
Figure~\ref{fig:abstraction_modalities} shows that both tested models pass through an intermediate
regime of abstraction before verbatim reconstruction. Strikingly, in CogVideoX the memory
trajectory reveals scene structure (street/beach) and characteristic motion (the direction of
the car or clouds) among the first properties to be acquired, while instance-specific appearance
still varies across initial noise realizations. Only later do the generated clips converge toward
the target. In FLUX.2, identity and pose likewise emerge before the exact appearance of the
training image. In Appendix~\ref{app:LoRA}, we further show that LoRA adaptation~\citep{hu2021loralowrankadaptationlarge} of SD~2.1 to a
single image produces a similar abstraction phase to that in Fig.~\ref{fig:trajectory}.
Together, these results suggest that abstraction before verbatim recall is not tied to the
SD~2.1 cross-attention interface or to our particular memory parameterization, and can persist
across changes in architecture, generative objective, and modality.
\vspace{-5pt}

\section{Limitations}
\label{sec:limitations}
\vspace{-5pt}
We identify four limitations.
(i) The observed semantic-first organization cannot be fully disentangled from the influence of text conditioning during pretraining. Although memory formation itself receives no textual supervision, data structure and language may both be required to shape the semantic directions along which abstraction emerges.
(ii) Our experiments on strongly out-of-distribution targets are limited. In these extreme regimes, the memorization trajectory changes qualitatively, and the abstraction phase narrows or disappears altogether.
(iii) The theoretical framework serves primarily as an interpretative model rather than a complete mechanistic account of the observed dynamics. Extracting more detailed predictions would require additional assumptions on the model structure, reducing the generality of the framework.
(iv) The fidelity--functionality trade-off is established for our memorization protocol and a limited set of functional probes. Different objectives or memory parameterizations may weaken, or completely avoid this trade-off.

\section{Conclusions}
\vspace{-5pt}
Results suggest that gist-like and verbatim recall can emerge at different stages of the same memorization process, as a datum is learned relative to a generative prior. What is worth storing first is not intrinsic to the datum, but depends on what the prior can reconstruct and which residual directions can be acquired most efficiently. The results suggest a downside of exact recall: increasing fidelity enriches an abstract memory, but can remove the invariances that make that memory reusable.

Several questions follow from this picture. The alignment between abstraction and semantic structure may reflect both language supervision and regularities already present in the data. Disentangling these contributions will require priors learned without text or under controlled data statistics. 
More broadly, the fidelity--functionality trade-off suggests that useful artificial memories need not be optimized all the way to exact reconstruction. Instead, their level of specificity could be adapted to the intended use: abstract memories may be preferable when flexibility, retrieval, or integration with new information matters, while more precise memories are needed for faithful reconstruction. Understanding and controlling this balance could inform the design of memory systems built on pretrained generative models. Finally, following what is acquired along a memorization trajectory may provide a way to expose the relevance structure of a generative prior itself.

\subsection*{AI use statement}
\vspace{-10pt}
LLMs implemented code, executed experiments, and substantially assisted results analysis and manuscript preparation, including organization, prose revision, and clarity checks. The authors reviewed and approved all scientific claims, experimental design, analysis, and final text.

\bibliography{iclr2027_conference}
\bibliographystyle{iclr2027_conference}

\clearpage
\appendix

\startcontents[appendices]

\section*{Appendix Overview}
\printcontents[appendices]{}{1}{}

\section{Extended related works}\label{app:related}

\paragraph{Generative priors and lossy compression.}
Viewing memory as storing what cannot be supplied by a prior is closely related to generative approaches to lossy compression. Classical rate--distortion theory formalizes the trade-off between representation rate and reconstruction accuracy, while the perception--distortion and rate--distortion--perception frameworks distinguish sample-wise fidelity from the requirement that reconstructions remain perceptually consistent with the data distribution~\citep{blau2018perception,blau2019rethinking,niu2025rate}. Close to our setting, \citet{tschannen2018deep} study distribution-preserving lossy compression, whose limiting regimes range from generation from the learned distribution at zero rate to exact reconstruction at sufficiently high rate. More recently, pretrained diffusion models have been used directly as strong priors for recovering information discarded by compression~\citep{relic2024lossy}. Information-theoretic work also studies representations that can be progressively refined while varying distortion and perceptual constraints~\citep{zhang2021universal}. \citet{bachmann2025flextok} and \citet{atanov2026videoflextok} learn ordered representations in which short token prefixes capture abstract content and additional tokens refine reconstruction. These works vary an explicit representation budget, such as bitrate or token-prefix length. Here, instead, representation size is fixed and the ordering unfolds over optimization time.

\paragraph{Compression and reconstructive memory.}
A parallel literature in cognitive science and computational neuroscience treats memory as a compressed representation whose missing information can be reconstructed from prior knowledge. From a Bayesian viewpoint, recall can be formalized as an interaction between noisy episodic evidence and prior knowledge at different levels of abstraction~\citep{hemmer2009bayesian}, while rate--distortion approaches have argued that limits on perceptual and memory resources naturally produce selective, lossy representations~\citep{bates2020efficient}. In the semantic-compression framework of \citet{nagy2020optimal}, compression through a learned generative model produces systematic memory distortions and a continuum between detailed and increasingly gist-like recall. \citet{fayyaz2022model} similarly propose that episodic memory stores an incomplete ``episodic gist'', with a generative semantic system completing information omitted from the trace. These works share with ours the idea that what must be explicitly stored depends on what a model of prior experience can reconstruct.

\paragraph{Generative memory and residual information.}
Recent work makes the relation between memory and generative prediction increasingly explicit. The hippocampo--neocortical model of \citet{spens2026hippocampo} describes hippocampal traces as compressed representations used to retrieve and condition a generative network, avoiding explicit storage of information that can instead be supplied by neocortical knowledge. The adaptive-compression framework of \citet{nagy2025adaptive} extends this idea to a setting in which semantic knowledge itself changes over time: surprising experiences can update the model rather than merely being compressed relative to a fixed one. Our construction deliberately freezes the generative prior, isolating the complementary problem of forming a single new memory relative to an existing body of knowledge.

A complementary perspective comes from \citet{chandra2025episodic}, who study how pre-existing hippocampal spatial structure changes what must be stored explicitly for episodic and associative recall. Their spatial scaffolds reduce sequence learning to structured transitions, with recall detail depending on available representational resources. More directly related to our residual formulation, \citet{lin2024images} show that reconstruction residuals of compressed visual representations predict both the strength of human memory traces and retrieval latency. This supports the broader view that information poorly accounted for by an existing representation can receive preferential memory resources.

\paragraph{Gist, meaning, and abstraction.}
Beyond the gist--verbatim distinction discussed in Sec.~\ref{sec:func-fidel}, a recent work in Statistical Mechanics formulates semantic compression as minimizing description length while preserving position in a semantic representation space, yielding distinct lossless/lossy and extractive/abstractive regimes~\citep{can2025statistical}. Although this work concerns semantic compression rather than memory formation, it provides a complementary formalization of abstraction as preserving semantic structure while discarding finer information.

\paragraph{Tilting and control of generative priors.}
Our theoretical description of memory as a learned modification of a frozen generative distribution connects to recent control-theoretic views of diffusion and flow models. \citet{uehara2024fine} formulate diffusion fine-tuning as entropy-regularized stochastic control relative to a pretrained process, while \citet{domingo2025adjoint} cast reward fine-tuning of diffusion and flow models as a stochastic optimal-control problem. More recent work directly considers target distributions obtained by exponentially tilting a base generative density~\citep{potaptchik2026tilt,domingo2026unified}. As opposed to this approach, our exponential-tilt model is not to be intended as part of a new fine-tuning algorithm, and we instead use it to build a minimal description of memory formation.

\paragraph{Learned conditioning and inversion.}
Beyond the personalization methods discussed in the main text, ELITE and IP-Adapter learn mappings from reference images to conditioning representations~\citep{wei2023elite,ye2023ipadapter}, while Null-text Inversion optimizes unconditional embeddings around an image-specific inversion trajectory to support reconstruction and text-guided editing~\citep{mokady2022nulltext}. These methods are concerned primarily with obtaining useful conditioning or editable reconstructions, while we focus on how the properties of a single learned conditioning signal change over the course of its optimization.

\section{Mathematical framework derivation}\label{app:derivation}
In this appendix we will provide a derivation for the equations presented in Section \ref{sec:framework}.\\
 Let us start by defining a forward noising process in $\mathbb{R}^d$ which gradually mixes a data point $\bm x_0 \sim p_0$ with Gaussian noise, as in \eqref{eq:forwardxt}:
 \begin{equation}
  \bm x_t = \alpha_t\bm  x_0 + \sigma_t \bm \epsilon\,, \quad \bm \epsilon \sim \mathcal{N}(0,\bm I)\,, \quad t\in[0,T]\,,
\end{equation}
as stated in Section \ref{sec:method}, one usually takes $\alpha_t\to 0$ and $\sigma_t \to 1$ as $t\to T$ so that the ending distribution $p_T$ is a standard Gaussian. At each timestep, the marginal distribution of the noised points $x_t$ is given by
\begin{equation}\label{eq:forwardpt}
p_t(\bm x)
=
\int d\bm x_0\,
p_0(\bm x_0)\,
\mathcal{N}
\!\left(
\bm x;\alpha_t \bm x_0,\sigma_t^2\bm I
\right)\,.
\end{equation}
Sampling from $p_0$ is done by inverting the forward process, i.e. by finding a process which, starting from $p_T$, brings us to $p_0$. The idea is that $p_T$ is easy to sample from, so getting samples from $p_0$ becomes as simple as drawing from $p_T$ and then following the backward process. This can be achieved in many different ways: we can for example train a model to infer the original $\bm x_0$ from $\bm x_t$, or to deduce the added noise $\bm \epsilon$, and then invert \eqref{eq:forwardxt}. These approaches are all related to the computation of the score function $\bm s(\bm x,t)=\nabla_{\bm x}\log p_t(\bm x)$ by Tweedie's formula \citep{efron2011tweedie}, which can be retrieved through direct computation from \eqref{eq:forwardpt}:
\begin{equation}
  \bm s_0(\bm x,t)=\nabla_{\bm x}\log\int d\bm x_0\,
p_0(\bm x_0)\,
\mathcal{N}
\!\left(
\bm x;\alpha_t \bm x_0,\sigma_t^2\bm I
\right)=\frac{\alpha_t \mathbb{E}[\bm x_0|\bm x_t]-\bm x_t}{\sigma_t^2}=-\frac{1}{\sigma_t}\mathbb{E}[\bm \epsilon|\bm x_t]\,.
\end{equation}
We now move on to the exponential tilt of the binary features $\bm \phi$ introduced in \eqref{eq:feature_tilt}:
\begin{equation*}
p_{\bm u}(\bm x_0)
=
\frac{
e^{\bm u^\top\bm\phi(\bm x_0)}\bm 
}{
Z(\bm u)
}
p_0(\bm x_0),
\qquad
Z(\bm u)
=
\mathbb{E}_{p_0}
\!\left[
e^{\bm u^\top\bm\phi(\bm X_0)}
\right]\,.
\end{equation*}
When noising samples from this new target distribution, we get a slightly modified version of the marginals in \eqref{eq:forwardpt}:
\begin{align}
  p_t^{\bm u}(\bm x_t)&= \frac{1}{Z(\bm u)}\int d\bm x_0\,
p_0(\bm x_0)\,
\mathcal{N}
\!\left(
\bm x;\alpha_t \bm x_0,\sigma_t^2\bm I
\right)\,e^{\bm u^\top\bm\phi(\bm X_0)}=\\
&=\frac{1}{Z(\bm u)}p_t(\bm x_t)\mathbb{E}_0
\!\left[
e^{\bm u^\top\bm\phi(\bm X_0)}
\,\middle|\,
\bm X_t=\bm x
\right]\label{eq:newpu}
\end{align}
where we have used Bayes' theorem to write
\begin{equation}
  p_0(\bm x_0)\,
\mathcal{N}
\!\left(
\bm x;\alpha_t \bm x_0,\sigma_t^2\bm I
\right)=p_0(\bm x_0)p(\bm x_t|\bm x_0)=p(\bm x_0|\bm x_t)p_t(\bm x_t)\,,
\end{equation}
and we have denoted by $\mathbb{E}_0[\cdot]$ the expectation over $p(\bm x_0|\bm x_t)$, i.e. the posterior induced by the original $p_0$ and the forward noising process.
The expression of \eqref{eq:newpu} allows us to directly derive the new score's decomposition \(\bm s_{\bm u}(\bm x,t)=\bm s_0(\bm x,t)+\delta\bm s_{\bm u}(\bm x,t)\) given by \eqref{eq:controlformula} \citep{kappen2005linear}:
\begin{equation}
\delta\bm s_{\bm u}(\bm x,t)
=
\nabla_{\bm x}
\log
\mathbb{E}_0
\!\left[
e^{\bm u^\top\bm\phi(\bm X_0)}
\,\middle|\,
\bm X_t=\bm x
\right]\,.
\end{equation}
The conditioning vector $\bm u$ is learned by choosing a single datum $\bm x^\star$ and making $p_u(\bm x_0)$ collapse on $p^\star(\bm x_0)=\delta(\bm x_0-\bm x^\star)$, i.e. by forcing the conditioned model to only produce $\bm x^\star$. This can be done by minimizing the weighted objective in \eqref{eq:memory_objective}, which we can write more explicitly as:
\begin{equation}
  \mathcal{L}(u)=\int_0^Td t\, w(t)\,\mathbb{E}_{p^\star}\,\left[||{\bm s_{\bm u}(\bm X_t, t)-\bm s^\star(\bm X_t, t)}||^2\right]\,,
\end{equation}
where
\begin{equation}
  \bm p^\star(\bm x_t,t) = \mathcal{N}
\!\left(
\bm x;\alpha_t \bm x^\star,\sigma_t^2\bm I
\right)
\end{equation}
is the marginal of the forward process starting from $\bm x^\star$, while 
\begin{equation*}
  \bm s^\star(\bm x_t,t) = -\sigma_t^{-2}(\bm x_t- \alpha_t \bm x^\star)
\end{equation*}
is the score associated to it, as found in \eqref{eq:scorestar}.\\
To further study the evolution of $\bm u$ induced by the gradient flow, we put ourselves in the SDE-based generation framework. In particular, we use the Ornstein-Uhlenbeck process in the noising phase, which corresponds to the variance preserving (VP) scheme:
\begin{equation}
  d\bm x_\theta=-\bm xd\theta+\sqrt{2}d\bm W_\theta
\end{equation}
where $d\bm W_\theta$ is a Wiener process and $\theta\in[0,\Theta]$. It can be shown that the solution of this process is given by:
\begin{equation}\label{eq:forwardsde}
  \bm x_t = e^{-\theta}\bm x_0 + \sqrt{1-e^{-2\theta}}\bm \epsilon
\end{equation}
which can be reduced to \eqref{eq:forwardxt} by putting $\theta=-\log\alpha_t$, $\sigma_t=\sqrt{1-\alpha_t^2}$ and taking $\Theta\to\infty$. If we now define a backward time $s=\Theta-\theta$ with negative increments $ds$, the backward process can be written as:
\begin{equation}\label{eq:backwardSDE}
  d\bm x_s = (-\bm x_s - 2\bm s(\bm x_s,\theta=\Theta-s))ds + \sqrt{2}d\tilde{\bm W}_s\,,
\end{equation}
where $d\tilde{\bm W}_s$ is independent from the original one.
In this framework, there is an elegant way of rewriting the objective $\mathcal{L}(\bm u)$: first, similarly to \citep{Lyu2012interpretation}, we can show that the Kullback–Leibler divergence between two distributions $p^1$ and $p^2$ noised through \eqref{eq:forwardsde} will evolve as 
\begin{align}
    \frac{d}{d\theta}D_{KL}(p_\theta^1||p_\theta^2)&=-\mathbb{E}_{p_\theta^1}\left[||\nabla_{\bm x}\log p_\theta^{1}(\bm X_\theta)-\nabla_{\bm x}\log p_\theta^{2}(\bm X_\theta)||^2\right]\\&=-\mathbb{E}_{p_\theta^1}\left[||\bm s^1(\bm X_\theta,\theta=\Theta-s)-\bm s^2(\bm X_\theta,\theta=\Theta-s)||^2\right]\,.
\end{align}
Where the minus sign is consistent with the fact that the two distributions are both converging toward a standard Gaussian. In the limit of large $\Theta$, $p_\Theta^1=p_\Theta^2=\mathcal{N}(0,\bm I)$: starting from this distribution and using the exact scores, the backward process (\eqref{eq:backwardSDE}) will reproduce the exact marginals $p_\theta^1$ and $p_\theta^2$ at $s=\Theta-\theta$. As a result, we can integrate the previous equation (with a sign flip) starting from $D_{KL}(p_\Theta^1||p_\Theta^2)=0$ to get
\begin{equation}
  D_{KL}(p^1||p^2)=\int_{0}^{\Theta}ds\, \mathbb{E}_{p_s^1}\left[||\bm s_1(\bm X_s,\theta=\Theta-s)-\bm s_2(\bm X_s,\theta=\Theta-s)||^2\right]\,,
\end{equation}
and then we can plug this expression, together with the likelihood time weighting $w(t)dt\propto ds$ \citep{song2021maximum} to get:
\begin{equation}\label{eq:scoredkl}
  \mathcal{L}(u)=D_{KL}(p^\star||p_{\bm u}) = -\log p_{\bm u}(\bm x^\star) =
-\bm u^\top\bm\phi(\bm x^\star) + \log Z(\bm u) + \mathrm{const}.
\end{equation}
We stress the fact that while this choice of $w(t)$ is the most convenient in this theoretical setting, it is not the one we use in the experiments. The expression  in \eqref{eq:scoredkl} allows us to easily get the gradient flow dynamics of $\bm u$ with learning rate $\eta$ found in \eqref{eq:gradientflow}:
\begin{equation}
  \frac{d\bm u}{d\tau}
=
-\eta\nabla_{\bm u}\mathcal{L}
=
\eta\,\bm r_{\bm\phi}(\bm u),
\end{equation}
where we have used the definition in \eqref{eq:feature_residual} 
\begin{equation*}
\bm r_{\bm\phi}(\bm u)
=
\bm\phi(\bm x^\star)
-
\mathbb{E}_{p_{\bm u}}
\left[
\bm\phi(\bm X)
\right]\,,
\end{equation*}
and we have used the fact that
\begin{equation}
  \nabla \log Z(\bm u) =\frac 1 {Z(\bm u)}
\int d\bm x_0 p(\bm x_0)\,\bm\phi(\bm x_0)\,
e^{\bm u^\top\bm\phi(\bm x_0)}=\mathbb{E}_{p_{\bm u}}
\!\left[
\bm\phi(\bm X_0)
\right]\,.
\end{equation}
We are only left with finding the derivative of $\bm r_{\bm\phi}(\bm u)$, which, by the chain rule, can be expressed as 
\begin{equation}
  \frac{d\bm r}{d\tau}=\frac{\partial\bm r}{\partial\bm u}\frac{d\bm u}{d\tau}\,,
\end{equation}
where $\frac{\partial\bm r}{\partial\bm u}$ is the matrix of entries $\frac{\partial r_i}{\partial u_j}$. These elements can be easily expressed in the following way:
\begin{align}
  \frac{\partial r_i}{\partial u_j}&=-\frac{\partial \mathbb{E}_{p_{\bm u}}
\!\left[
\bm\phi_i(\bm X_0)
\right]}{\partial u_j}=-\frac{\partial }{\partial u_j}\frac 1 {Z(\bm u)}
\int d\bm x_0 p(\bm x_0)\,\bm\phi(\bm X_0)\,
e^{\bm u^\top\bm\phi(\bm X_0)}=\\&=-\mathbb{E}_{p_{\bm u}}
\!\left[
\bm\phi_i(\bm X_0)\phi_j(\bm X_0)
\right]+\mathbb{E}_{p_{\bm u}}
\!\left[
\bm\phi_i(\bm X_0)
\right]\mathbb{E}_{p_{\bm u}}
\!\left[
\bm\phi_j(\bm X_0)
\right]=-[\bm \Sigma_{\bm u}]_{ij}\,,
\end{align}
which finally gets us to the other expression in \eqref{eq:gradientflow}:
\begin{equation}
\frac{d\bm r_{\bm\phi}}{d\tau}
=
-\eta\bm\Sigma_{\bm u}\bm r_{\bm\phi},
\end{equation}

\section{Implementation details}\label{app:implementation}

\paragraph{Memory parameterization.} The memory is a set of $M$ learnable vectors
$\bm C=\{\bm c_1,\ldots,\bm c_M\}$, $\bm c_m\in\mathbb{R}^{1024}$, living in the
cross-attention context space of the frozen UNet ($8{,}192$ trainable parameters at
$M{=}8$). At every cross-attention layer the memory is read through a second attention
pass that reuses the layer's frozen query/key/value/output projections. The value and output projections are applied without their biases, so a zero memory's contribution is exactly equal to zero. As a consequence, at $\bm C=\bm 0$ generation coincides with the empty-caption model exactly. Notice that this parallel construction allows us to vary freely the size of the memory $M$, instead of being tied to the text embedding space size (77 tokens). The text pathway is conditioned on the empty caption throughout the whole process.

\paragraph{Initialization.} Each vector is drawn i.i.d.\ Gaussian with standard deviation
$0.01$ (per-vector norm ${\approx}0.32$). We use this random initialization instead of putting all vectors to zero because the latter is degenerate: all
keys would coincide and receive identical gradients indefinitely.

\paragraph{Training.} We minimize the frozen model's own denoising objective
(\eqref{eq:memory_objective}) on the target latent (VAE posterior mean of $\bm x^\star$ at
$768$px), with the native $v$-prediction target and $t$ sampled uniformly. Optimization
uses AdamW with learning rate $10^{-2}$, no weight decay, and a \emph{constant} schedule
with no warm-up. Each gradient step averages $8$ independent $(t,\bm\epsilon)$ draws against
the same target. Runs last $5{,}000$
steps with ${\sim}20$ checkpoints on a fixed grid unless otherwise specified.

\paragraph{Sampling.} All samples use DPM-Solver++ (multistep, order $3$), $50$ steps,
$768$px, classifier-free guidance $\omega=3$; the unconditional branch is the same model
with the memory detached. We use $\omega=3$ for the caption-conditioned generated images used as reference as well, so that memory- and caption-conditioned samples are always compared under an
identical protocol. Sampling is deterministic given the noise seed, which is what licenses
the seed-paired comparisons of Section~\ref{sec:analysis}. Evaluation seeds are fixed
per experiment and disjoint from any seed involved in creating a training target.

\paragraph{Compute.} Each single-image run (training plus checkpoint sampling) takes a few
hours on one A100/H100-class GPU; all experiments in the paper are single-GPU.

\section{Experimental details}\label{app:experiments}

\subsection{From gist to verbatim (Fig.~\ref{fig:trajectory})}\label{app:exp_trajectory}

\paragraph{Targets.} The bank contains $30$ in-distribution targets ($10$ animals, $10$
vehicles, $10$ faces), each generated by the frozen model itself at $\omega=7.5$, $50$
steps, $768$px, from a per-image caption with a recorded generating seed. 
All $30$ are trained at $M\in\{1,8,32,64\}$.

\paragraph{Evaluation samples.} At each of ${\sim}20$ checkpoints we store $S=20$ samples
from fixed held-out seeds under the protocol of Appendix~\ref{app:implementation}.
Because targets are model generated, a target's own generating seed can fall inside the
evaluation set. In that case, the generating seed is excluded and the run is evaluated on the remaining
$19$ seeds, so no evaluation sample ever shares the initial noise that produced its
run's target. 

\paragraph{Bias and variance.} For an embedding $E$ and samples
$\{\bm g_s\}_{s=1}^{S}$ at one checkpoint, let $\bar{\bm e}=S^{-1}\sum_s E(\bm g_s)$. We
study
\begin{equation}
    \operatorname{bias}^2
    =
    \frac{\left\|\bar{\bm e}-E(\bm x^\star)\right\|^2}{\left\|E(\bm x^\star)\right\|^2},
    \qquad
    \operatorname{variance}
    =
    \frac{1}{S}\sum_s
    \frac{\left\|E(\bm g_s)-\bar{\bm e}\right\|^2}{\left\|E(\bm x^\star)\right\|^2},
    \label{eq:bias_variance}
\end{equation}
 which together form an exact decomposition of the normalized mean squared distance to the target embedding. The semantic space is the DINOv2  \citep{oquab2024dinov2learningrobustvisual}embedding (ViT-B/14), mean-pooled over patch tokens and unit-normalized, while the surface space is the raw VAE latent.

\paragraph{Half-decay steps.} We define the half-decay step of a curve as the training step at which it has
covered half of its total fall. We compute it through a linear interpolation of the checkpoints immediately before and after the crossing. This way, we assume no functional form for the curve.
In Section~\ref{sec:gist_to_verbatim}, we have computed the per-image ratio
between the half-decay steps of the surface and semantic bias$^2$ curves. The value quoted in the text is the median of the distribution computed over the $30$
images.

\subsection{Generalization through the prior (Fig.~\ref{fig:generalisation}a,b)}\label{app:exp_generalisation}

\paragraph{Setup.} Each caption defines a reference pool: $100$ generated images at fixed noise
seeds under the sampling protocol of Appendix~\ref{app:implementation}, split into 50 train and 50 held-out test seeds. A memory is trained from scratch on
$N$ of the training pictures ($N{=}1$ or $N{=}50$, $M{=}8$, $4{,}000$ steps, $18$
checkpoints with held-out sampling at every checkpoint). Twelve face captions enter the
pooled result, one memory per caption at each $N$, all under one protocol: target
images, memory samples, and reference pools share the same sampler, resolution, and
$\omega=3$. Again, training seeds never intersect the test split, so no reference image is ever
evaluated against what was trained on.

\paragraph{Paired advantage.} At a checkpoint, let $\bm g_s$ be the memory's sample at
held-out seed $s$ and $\bm y_s$ the caption's own generated image from that same noise. Calling
$\hat E$ the unit-normalized DINOv2 embedding, we compute the embedding distances
$D_{ss'}=\lVert\hat E(\bm g_s)-\hat E(\bm y_{s'})\rVert$ over the $S=50$ test seeds and then
define
\begin{equation}
\mathrm{adv}
\;=\;
\frac{1}{\bar D}\,\frac{1}{S}\sum_s
\Big(\tfrac{1}{S-1}\textstyle\sum_{s'\neq s} D_{ss'}\;-\;D_{ss}\Big),
\label{eq:advantage}
\end{equation}
which corresponds to the mean excess distance of a sample to \emph{other} seeds' outputs over its own seed's,
normalized by the mean entry $\bar D$ so values are comparable across captions. The
statistic is zero both for an untrained memory (samples unrelated to any seed) and for a
fully memorized one (identical samples at every seed), and it can only be positive when the
output varies with the seed the way the caption's generated images do. 

\paragraph{Significance and controls.} Holding $D$ fixed, we randomly re-pair samples
with references and recompute \eqref{eq:advantage}.
 As a
negative control, the same samples are scored against pools generated from three
unrelated captions at the same seeds, leading to a metric which stays close to 0 throughout the training process.

\subsection{Positional specificity of addressing (Fig.~\ref{fig:generalisation}c)}\label{app:exp_positional}

\paragraph{Probe.} In the positional addressing experiment, we evaluate each memory using a fixed ensemble of $P{=}12$ held-out probe images generated from 12 different captions, for all three classes: animals, vehicles, and faces. The probes belong to the same class as the memory's training target, and no probe image is a training target in any run. To compute the metrics, we add noise to the probes at timestep $t{=}800$ using a fixed noise draw and pass them through the UNet with the memory being evaluated attached. This evaluation noise is separate from the noise used to generate the probe images.

\paragraph{Measurement.} At every cross-attention layer we record the query--key cosine similarity
$a_{p,\ell,m}=\cos(\bm q_{p,\ell},\bm k_m)$ for probe image $p$, position $\ell$, and
memory token $m$, averaged over heads. Fixing a layer and a token and writing
$\bar a_\ell=\frac1P\sum_p a_{p,\ell}$ for the image-averaged map, we define
\begin{equation}
V_{\mathrm{pos}}
=
\operatorname{Var}_\ell\!\left(\bar a_\ell\right),
\qquad
V_{\mathrm{cont}}
=
\mathbb{E}_\ell\!\left[\operatorname{Var}_p\!\left(a_{p,\ell}\right)\right],
\label{eq:vpos_vsem}
\end{equation}
so that $\operatorname{Var}_{p,\ell}(a)=V_{\mathrm{pos}}+V_{\mathrm{cont}}$ exactly, by
the law of total variance: $V_{\mathrm{pos}}$ is the positional structure shared by all
probe images, $V_{\mathrm{cont}}$ the residual across-image variability. Positional
structure that differs per image is counted in $V_{\mathrm{cont}}$, so the reported
fraction $f=V_{\mathrm{pos}}/(V_{\mathrm{pos}}+V_{\mathrm{cont}})$ is a conservative
estimate of positional addressing. Components are averaged over tokens and the $16$
layers before the ratio, so layers contribute in proportion to the variance they carry.

\paragraph{Runs and baseline.} We measure $30$ memories per memory size ($10$ per class) at
$M\in\{1,8,32,64\}$ at every stored checkpoint, and plot their mean $\pm$ s.e.m.\,. With a
single token the softmax weight is identically one, so the $M{=}1$ addressing pathway
receives no gradient; its slow drift reflects the token moving under the value-pathway
loss, read through the frozen key projection, and serves as the baseline against which
the $M\geq 8$ rise is read.

\subsection{Editability (Fig.~\ref{fig:editability})}
\label{app:exp_edit}

\paragraph{Data.}
We use the ten model-generated face targets described in Appendix~\ref{app:exp_trajectory} and their existing single-image memories. We denote each target by $x_{\mathrm{orig}}$ (corresponding to $\bm x^\star$ in Section~\ref{sec:method}) and its memory after $s$ optimization steps by $\bm C^{(s)}$. For each target, we generate two edited counterparts $x_{\mathrm{edit}}$ using FLUX.2 Kontext~\citep{batifol2025flux1kontext}, yielding $20$ same-identity target--edit pairs. Edits vary pose, expression, accessories, and scene or background. The qualitative panel additionally includes two cross-identity examples in which the edited input depicts a different person from the memory's stored subject.

For both editability and one-shot recognition (Appendix~\ref{app:exp_recognition}), we evaluate the randomly initialized memories ($s=0$) and checkpoints $s\in\{50,200,300,400,600,800,1200,2000,3000,5000\}$. The initialization checkpoint retains the memory branch and is distinct from the memory-free prior, for which that branch is disabled.

\paragraph{U-turn protocol.}
Following diffusion-based editing~\citep{meng2021sdedit}, we partially noise the edited image and complete denoising under the target's memory. We encode the edit as $z_0^{\mathrm{edit}}=\mathcal{E}(x_{\mathrm{edit}})$ using the frozen VAE encoder and sample
\begin{equation}
z_{t_{\mathrm{switch}}}
= \alpha_{t_{\mathrm{switch}}}z_0^{\mathrm{edit}}
+ \sigma_{t_{\mathrm{switch}}}\varepsilon,
\qquad \varepsilon\sim\mathcal{N}(0,I).
\label{eq:app-sdedit-init}
\end{equation}
Here, $\alpha_t=\sqrt{\bar\alpha_t}$ and $\sigma_t=\sqrt{1-\bar\alpha_t}$ are the signal and noise coefficients of the SD2.1 schedule, consistent with Section~\ref{sec:method}. We then denoise from $t_{\mathrm{switch}}$ to $0$ with classifier-free guidance:
\begin{equation}
\hat v_t = v_\theta(z_t,t,\varnothing)
+ \omega\bigl[v_\theta(z_t,t,\bm C^{(s)}) - v_\theta(z_t,t,\varnothing)\bigr].
\label{eq:app-cfg}
\end{equation}
The frozen $v$-prediction denoiser $v_\theta$ receives empty-caption text conditioning $\bm c_{\varnothing}$ in both branches; $\varnothing$ denotes the memory-disabled branch. Thus, the memory is the sole additional conditioning signal.

We use stochastic DDIM sampling with $\eta=1$, $50$ sampling steps, guidance scale $\omega=3$, and resolution $768\times768$, with $t_{\mathrm{switch}}=800$ on the $T=1000$ diffusion-time grid. We write the output as $\hat x_k=\mathcal{U}(x_{\mathrm{edit}},\bm C^{(s)},t_{\mathrm{switch}},\xi_k)$, where $\xi_k$ seeds both the forward noise and the stochastic denoising trajectory. We generate $K=2$ samples per target--edit pair at each checkpoint.

\paragraph{Metric.}
Let $E_{\mathrm{DINO}}$ denote the unit-normalized DINOv2 \texttt{[CLS]} embedding (ViT-B/14, \texttt{dinov2-base}), and define the cosine distance $d_{\mathrm{DINO}}(a,b)=1-E_{\mathrm{DINO}}(a)^\top E_{\mathrm{DINO}}(b)$. For each target--edit pair and checkpoint, we compute
\begin{equation}
D_{\mathrm{edit}} = \frac{1}{K}\sum_{k=1}^{K} d_{\mathrm{DINO}}(\hat x_k,x_{\mathrm{edit}}),
\qquad
D_{\mathrm{orig}} = \frac{1}{K}\sum_{k=1}^{K} d_{\mathrm{DINO}}(\hat x_k,x_{\mathrm{orig}}).
\label{eq:app-dist}
\end{equation}
Lower $D_{\mathrm{edit}}$ indicates greater similarity to the edited input, whereas lower $D_{\mathrm{orig}}$ indicates greater similarity to the stored target. Increasing $D_{\mathrm{edit}}$ together with decreasing $D_{\mathrm{orig}}$ is consistent with the memory overwriting the edit as it approaches verbatim recall. At each checkpoint we report the mean of $D_{\mathrm{edit}}$ and $D_{\mathrm{orig}}$ over the
$N=20$ target--edit pairs.

\subsection{One-shot recognition (Fig.~\ref{fig:zeroshot})}
\label{app:exp_recognition}

\paragraph{Data.}
We use thirty face targets, corresponding to thirty different identities: the ten face targets and memory checkpoints of Appendix~\ref{app:exp_edit}, plus twenty additional identities whose memories are optimized with the identical protocol ($M=8$ tokens, same learning rate, batch size, resolution and $5{,}000$ steps). Using the same editing pipeline, we generate ten held-out images per identity, yielding $300$ images with variations in pose, expression, or scene, all of which are excluded from memory training. The task tests whether a memory learned from a single target image recognizes other realizations of the same identity.

\paragraph{Scoring protocol.}
We encode each held-out image $x$ as $z_0=\mathcal{E}(x)$ and draw $K=20$ uniformly sampled timestep--noise pairs $(t_k,\varepsilon_k)$, shared across candidate memories for that image. Each pair defines a noised latent and its $v$-prediction target:
\begin{equation}
z_{t_k} = \alpha_{t_k}z_0 + \sigma_{t_k}\varepsilon_k,
\qquad
v_{t_k} = \alpha_{t_k}\varepsilon_k - \sigma_{t_k}z_0,
\label{eq:app-oneshot-noise}
\end{equation}
where $\varepsilon_k\sim\mathcal{N}(0,I)$ and the encoder, denoiser, and noise coefficients are as defined in Appendix~\ref{app:exp_edit}. We score each memory by its estimated expected denoising error:
\begin{equation}
\hat L\bigl(x,\bm C_j^{(s)}\bigr)
= \frac{1}{K}\sum_{k=1}^{K}
\bigl\lVert v_\theta\bigl(z_{t_k},t_k,\bm C_j^{(s)}\bigr) - v_{t_k}\bigr\rVert_2^2\,,
\label{eq:app-oneshot-loss}
\end{equation}
where by $\bm C_j^{(s)}$ we denote the identity $j$'s memory at optimization step $s$.
The text pathway remains conditioned on the empty caption. We evaluate the denoiser directly, without a reverse diffusion trajectory, using images at resolution $768\times768$. All candidate memories are compared at the same optimization step.

\paragraph{Metric.}
We rank identities by increasing denoising error and assign each held-out image to
\begin{equation}
\hat{\jmath}^{(s)}(x)
= \operatorname*{arg\,min}_{j\in\mathcal{J}}
\hat L\bigl(x,\bm C_j^{(s)}\bigr)\,,
\label{eq:app-oneshot-argmin}
\end{equation}
where $\mathcal{J}$, with $|\mathcal{J}|=30$, denotes the set of all identity indices. Top-1 and top-3 accuracy measure the fraction of images whose true identity appears among the one or three lowest-error candidates, respectively, with chance levels $1/30$ and $0.1$. Ties are broken uniformly at random; this matters only at $s=0$, where all memories share the same initialization and accuracy equals chance. At each checkpoint, we compute each identity's own accuracy over its 10 held-out queries, then report the median and interquartile range (25th--75th percentile) of these $N=30$ per-identity accuracies. The qualitative examples in Fig.~\ref{fig:zeroshot} (top) rank the five queries against the original ten identities.

\subsection{Text--memory retrieval (Fig.~\ref{fig:retrieval})}
\label{app:exp_retrieval}

\paragraph{Data.}
We use a separate set of $N=15$ model-generated identities, distinct from those used for
editability and one-shot recognition. For each identity, we construct three text cues at
different levels of specificity and generate one target image per cue level with SD2.1
(Fig.~\ref{fig:retrieval}, top). We train a single memory on the \emph{detailed}-level image
alone, following Section~\ref{sec:method}, without supplying any text cue during training. The
other three generated images are not used for memory training. We denote identity $i$'s memory
after $s$ optimization steps by $\bm C_i^{(s)}$. All three cue levels are evaluated against this
same memory, using the initialization and evaluation checkpoints specified in
Appendix~\ref{app:exp_edit}.

\paragraph{Text cues.}
\emph{1 detail} adds one distinguishing attribute; and \emph{brief} and \emph{detailed} are automatically generated descriptions at two levels of detail. For one identity, the three cues are:
\begin{quote}
\emph{1 detail}: ``a photo of a man with short light brown hair''\\
\emph{brief}: ``A man standing in front of a colorful wall.''\\
\emph{detailed}: ``The image shows a man standing in front of a colorful wall with a smile on his face.''
\end{quote}

\paragraph{Text and memory representations.}
For cue level $\ell$, let $\bm t_i^{(\ell)}\in\mathbb{R}^{1024}$ denote the EOS-token representation produced by the frozen SD2.1 text encoder. We represent each memory by the mean of its $M=8$ conditioning vectors:
\[
\bar{\bm c}_i^{(s)}
= \frac{1}{M}\sum_{m=1}^{M}\bm c_{i,m}^{(s)}.
\]
Memory vectors and text-token representations share the same dimensionality and frozen UNet cross-attention key/value projections. We compare the pooled memory and text representations using cosine similarity.

\paragraph{Metric.}
We measure how well each identity's own text cue at level $\ell$ retrieves its own memory,
relative to the other $N-1$ identities' memories:
\begin{equation}
\mathrm{margin}_i^{(\ell)}(s)
= \cos\!\bigl(\bm t_i^{(\ell)},\bar{\bm c}_i^{(s)}\bigr)
- \frac{1}{N-1}\sum_{j\neq i}
\cos\!\bigl(\bm t_i^{(\ell)},\bar{\bm c}_j^{(s)}\bigr).
\label{eq:app-retrieval-margin}
\end{equation}
A positive margin indicates greater similarity to the identity's own memory than to the other
identities' memories on average. Cues shared by multiple identities remain in the comparison
set.

\subsection{Other models and modalities (Fig.~\ref{fig:abstraction_modalities})}
\label{app:exp_extensions}

\paragraph{Shared setup.}
We extend the memory construction of Section~\ref{sec:method} to CogVideoX-2b and FLUX.2. In both experiments, we freeze the pretrained model and optimize only $M=8$ conditioning vectors on a single target using the backbone's native training objective and empty-prompt text conditioning. The memory branch uses a separate softmax over memory keys and frozen projection weights with biases removed, ensuring zero contribution when the memory vectors are zero. We initialize the vectors with independent Gaussian entries of standard deviation $0.01$, keeping the initial model close to the empty-prompt prior. At generation, we apply classifier-free guidance with scale $\omega=3$, combining the memory-conditioned prediction with the zero-memory prediction, which is the empty-prompt prior.

\paragraph{CogVideoX-2b.}
CogVideoX-2b~\citep{yang2024cogvideox} is a latent video diffusion transformer with a causal 3D VAE and a native $v$-prediction objective. Each transformer block applies joint attention to text and spatiotemporal video tokens, using a hidden dimension of $1920$ with $30$ attention heads. Text conditioning is obtained from a T5 encoder and held at the empty caption throughout memory training and evaluation. We generate a single $49$-frame target video at resolution $480\times720$ using the frozen model and a hand-written caption describing a fixed-camera scene. The VAE encodes this target into a latent with $13$ temporal positions, $16$ channels, and spatial dimensions $60\times90$. The generating caption is not supplied during memory training. We represent the memory by eight $1920$-dimensional vectors and attach a memory branch to the joint attention in every transformer block. The joint text--video queries attend to memory keys and values obtained through the frozen key and value projections, with projection biases removed. We apply the base attention pathway's key normalization to the memory keys. The branch output passes through the frozen output projection without bias and is added to the joint attention output before it is separated into text and video streams. We train on the target latent using the native $v$-prediction parameterization of the denoising objective in~\eqref{eq:memory_objective}, sampling timesteps uniformly and drawing fresh noise at each update. Optimization uses AdamW with learning rate $10^{-2}$ and no weight decay for $3000$ steps. A $300$-step linear warm-up is followed by cosine decay to one tenth of the peak learning rate. We save checkpoints throughout training, with denser coverage during the first $500$ updates. 
Evaluation uses $50$ sampling steps, held-out noise seeds, and tiled VAE decoding. 

\paragraph{FLUX.2.}
We use the approximately four-billion-parameter base variant of FLUX.2~[klein]~\citep{flux2024}, a rectified-flow transformer. We attach the memory branch to its double-stream blocks, which jointly process text and image tokens, and leave the subsequent single-stream blocks unchanged. The eight memory vectors lie in the text-feature space consumed by the frozen context embedder. In each double-stream block, image queries attend to memory keys and values obtained by passing these vectors through the context embedder and the text-side key/value projections, with biases removed. We apply the corresponding key normalization and assign every memory token the first text rotary position. The memory branch output passes through the frozen image output projection without bias and is added to the image stream. For this backbone, we use the native flow-matching objective in place of the diffusion denoising objective. Given a packed target latent $\mathbf{z}_0$ and noise $\boldsymbol{\epsilon}\sim\mathcal{N}(\mathbf{0},\mathbf{I})$, the velocity target is $\mathbf{v}=\boldsymbol{\epsilon}-\mathbf{z}_0$. We sample timesteps from the flow-matching schedule with its image-sequence-dependent shift and average eight independent timestep-noise pairs per optimization step. Training uses AdamW with learning rate $10^{-2}$ and no weight decay. Empty-prompt conditioning is fixed throughout.
Sampling uses 50 flow-matching Euler steps at resolution $1024\times1024$. At each step, we combine memory-conditioned and zero-memory predictions using the shared guidance scale $\omega=3$.

\section{Additional experiments}\label{app:additional}

\subsection{Out-of-distribution targets}\label{app:add_ood}

\begin{figure}[t]
  \centering
  \includegraphics[width=\textwidth]{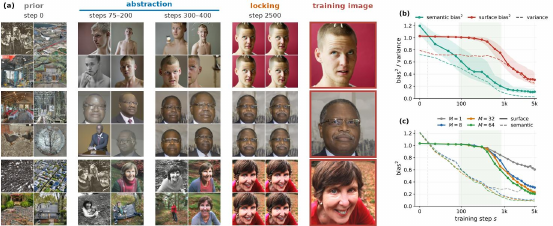}
  \caption{\textbf{Abstraction and locking on real photographs.} The
  external data counterpart of Fig.~\ref{fig:trajectory}, same layout.
  \textbf{(a)} three memories ($M{=}8$), each trained on a single FFHQ photograph (red
  frame) and sampled at held-out noise seeds: prior draws at initialization, other people
  sharing the target's type, then the target's identity with expression, lighting and
  background still varying, then verbatim recall. 
  \textbf{(b)} bias$^2$ and variance over all $80$ single-image memories at $M{=}8$
  ($20$ per class; median$\pm$IQR), semantic (DINOv2) and surface (VAE) spaces as in
  Fig.~\ref{fig:trajectory}b. \textbf{(c)} the same distances by memory size over the
  $40$ images trained at every $M$ ($10$ per class).}
  \label{fig:trajectory_ood}
  \vspace{-15pt}
\end{figure}

\paragraph{Setup.} Out-of-distribution targets are real photographs at $768$px: faces
from FFHQ~\citep{karras2019style}, and animals, vehicles, and plants from
DOCCI~\citep{onoe2024docci}, a corpus of photographs with detailed human-written
descriptions. We train $80$ single-image memories at $M{=}8$ ($20$ per class), $40$ of
which ($10$ per class) are additionally trained at every $M\in\{1,8,32,64\}$ for the
size comparison. These targets were never generated by the model, and no caption is
involved at any stage; training, sampling, and evaluation follow
Appendices~\ref{app:implementation} and~\ref{app:exp_trajectory} unchanged, except that
no target has a generating seed, so no seed exclusion applies.

\paragraph{The same two regimes, a narrower window.} Figure~\ref{fig:trajectory_ood}
reproduces the trajectory of Fig.~\ref{fig:trajectory} on photographs: the semantic bias$^2$
drops before the surface bias$^2$ later, and the diversity of the generated images collapses only when memorization is completed. The memory size $M$ is again important only for the final surface residual, rather than the for the timescale of early semantic acquisition. However, the separation between the two regimes contracts, and the per-image median ratio of surface to
semantic half-decay steps is $10.6$,
against $21.4$ for in-distribution targets. The contraction is visible in the grid
itself, whose abstraction columns sit at steps $75$--$400$ versus $75$--$600$ for
in-distribution targets. This phenomenon could be expected from the fact that a photograph's semantic content
is less readily expressible through the prior, so less of the image can be delegated to it. As we discuss in the Limitations, the same mechanism could lead to the disappearance of the abstraction phase for strongly out-of-distribution targets.

\paragraph{Addressing.} The positional-fraction transition of
Appendix~\ref{app:exp_positional} was nearly identical on out-of-distribution
memories ($20$ runs per size, animals and vehicles): a flat content-driven
plateau through abstraction, followed by a size-ordered rise as locking sets in,
above the $M{=}1$ drift baseline. Thus, the mechanism of
Section~\ref{sec:analysis} extends beyond model-generated targets.


\subsection{Strongly out-of-distribution targets}\label{app:add_ood_strong}

\begin{figure}[t]
  \centering
  \includegraphics[width=0.6\textwidth]{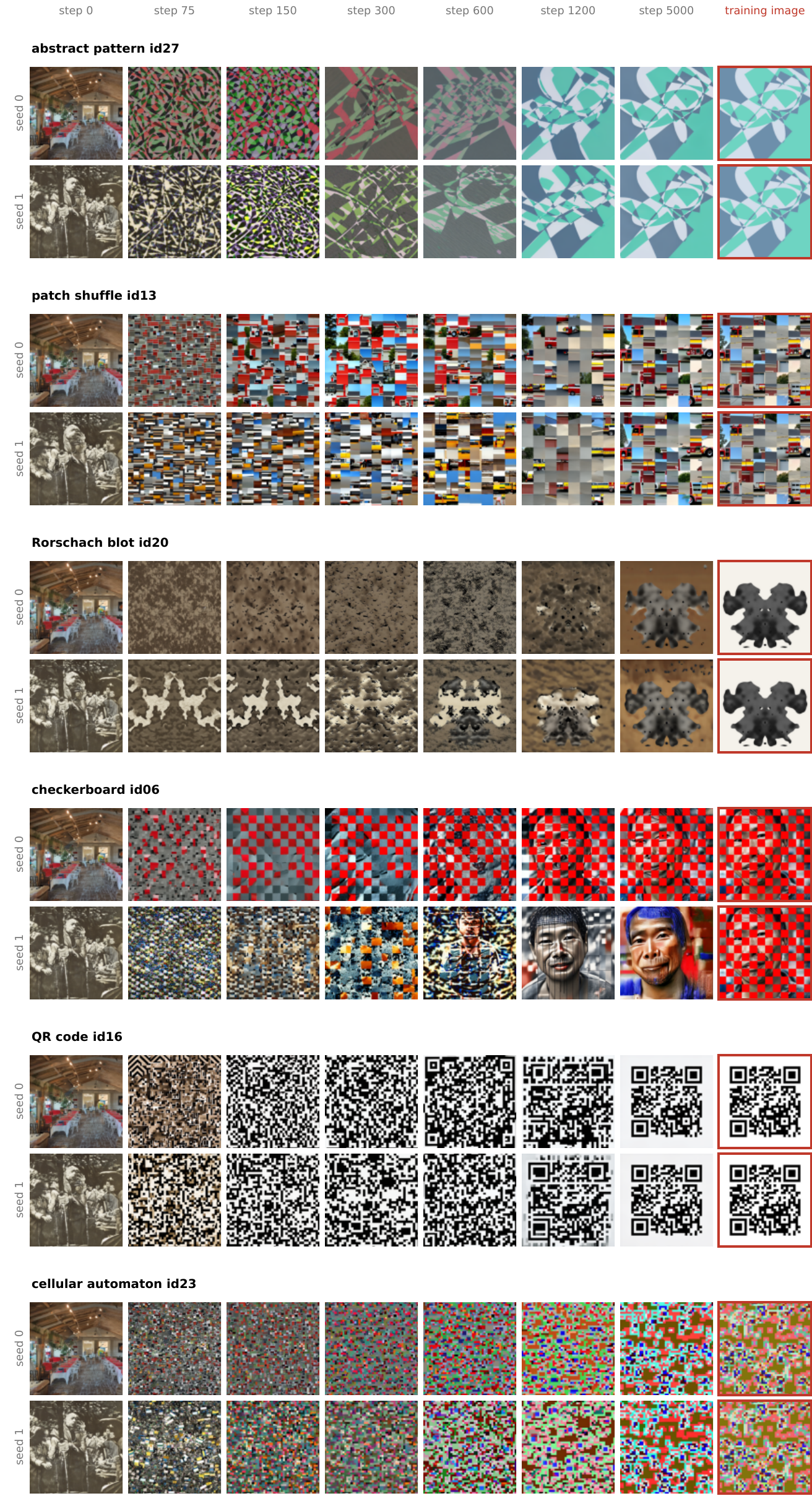}
  \caption{\textbf{Memorization trajectories, strongly out-of-distribution targets.} Same
  layout as Fig.~\ref{fig:qual_ood}, one memory ($M{=}8$) per class. Each class shows the run
  whose ratio of surface to semantic half-decay steps is the median of its class.}
  \label{fig:ood_strong_samples}
\end{figure}

\begin{figure}[t]
  \centering
  \includegraphics[width=\textwidth]{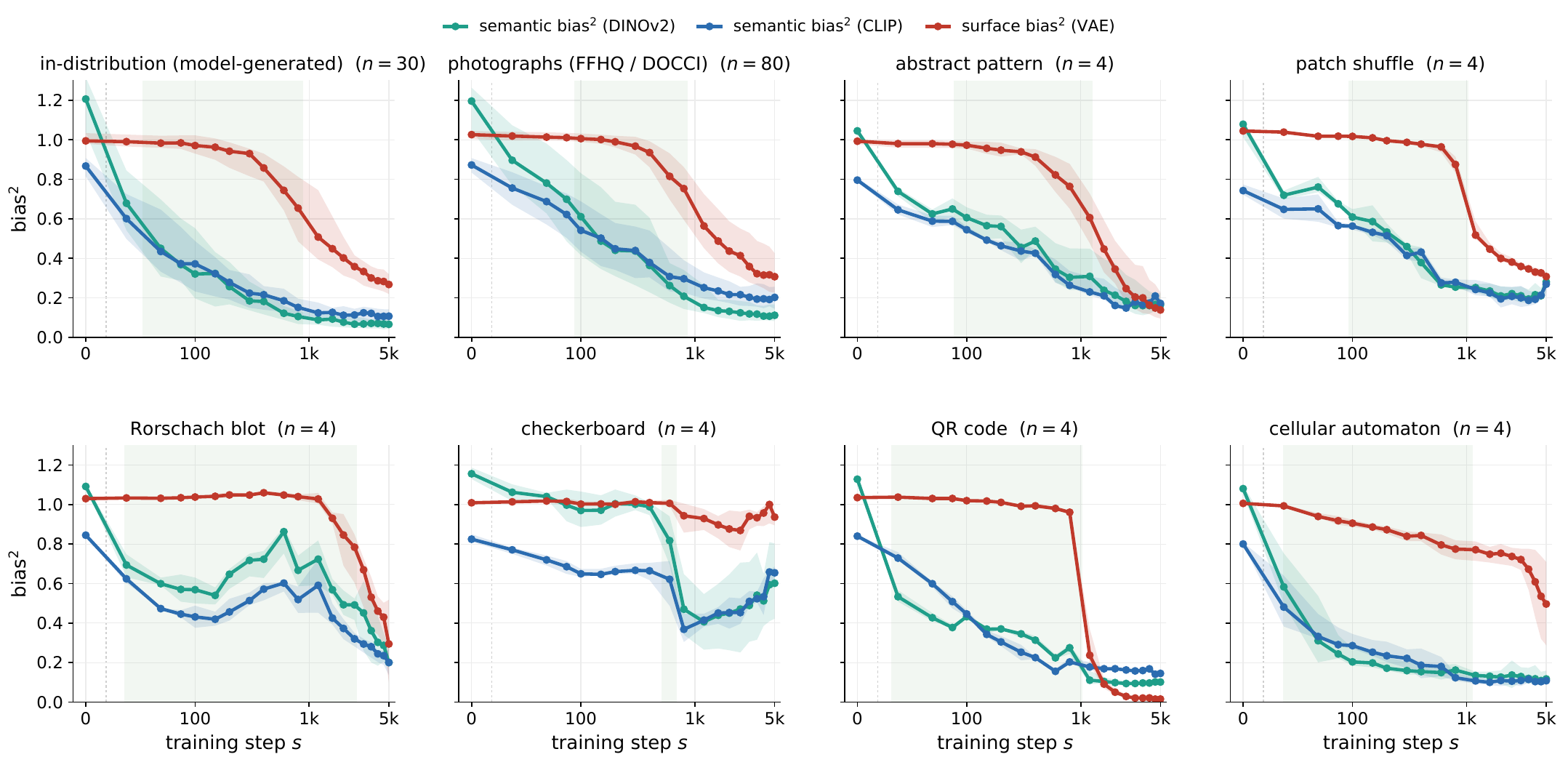}
  \caption{\textbf{Semantic and surface bias$^2$ on strongly out-of-distribution targets.}
  Median$\pm$IQR over the four targets of each class, in DINOv2 and CLIP (semantic) and VAE
  (surface) spaces; the in-distribution bank and the photographs of
  Appendix~\ref{app:add_ood} are shown for reference.}
  \label{fig:ood_strong_curves}
\end{figure}

\begin{figure}[t]
  \centering
  \includegraphics[width=\textwidth]{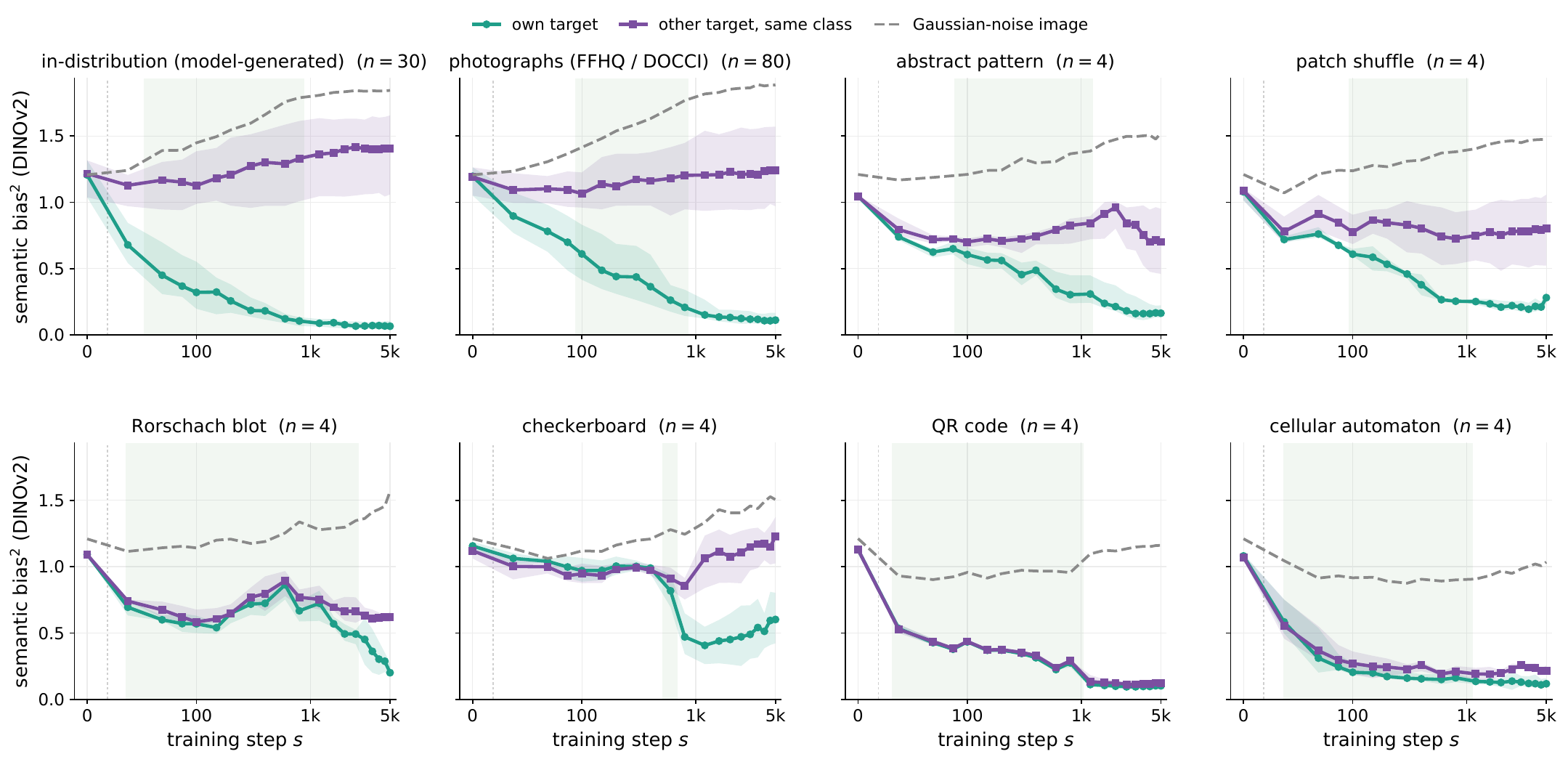}
  \caption{\textbf{Instance, class, or generic change.} DINOv2 bias$^2$ of the same samples
  measured against their own target, against another target of the same class, and against a
  fixed Gaussian-noise image. Only a gap between the first two curves indicates
  target-specific semantic information.}
  \label{fig:ood_strong_controls}
\end{figure}

\paragraph{Setup.} To probe limitation (ii) of Section~\ref{sec:limitations}, we built
targets designed to lie far from the model's training distribution, four in each of six
classes: abstract patterns made of random shapes and colours, photographs whose patches are
randomly permuted, Rorschach-like inkblots, images in which every other square is replaced by
pure red, QR codes with random content, and the states of random cellular automata
(Fig.~\ref{fig:ood_strong_samples}, last column). The classes differ in the way they are out
of distribution: inkblots and QR codes are genres the model knows but whose instances are
arbitrary, while the other classes are compositions that no familiar genre describes.
Training, sampling, and evaluation follow Appendices~\ref{app:implementation}
and~\ref{app:exp_trajectory} unchanged.

\paragraph{Separating instance from class.} On such targets, a decrease of the semantic
bias$^2$ is ambiguous: the samples may be acquiring the target, only its class, or simply
changing in a way that happens to reduce the distance. We therefore also measure the
distance of the same samples to another target of the same class, and to an image of
Gaussian noise (Fig.~\ref{fig:ood_strong_controls}). Information specific to the target is
present only when the distance to the target falls below the distance to the other member
of its class.

\paragraph{Abstraction depends on the target.} The trajectories split into three groups.
For abstract patterns and shuffled photographs, the abstraction regime survives: the
semantic bias$^2$ drops well before the surface bias$^2$, with a separation comparable to
that of in-distribution targets and photographs (Fig.~\ref{fig:ood_strong_curves}), and the
distance to the target soon falls below that to the other members of the class. The samples
first reproduce the palette and coarse layout of the target, and only later its exact
configuration. For inkblots, the genre is acquired almost immediately, but the samples remain
as close to other inkblots as to the target until the surface is fixed: the memory moves from
``an inkblot'' to ``this inkblot'' without an intermediate regime. Finally, the red
checkerboards are not stored within the training budget: the surface bias$^2$ barely
decreases, and samples tend to recover the underlying image while losing the tiling.

QR codes and cellular automata are memorized, and their samples suggest a genre-first
trajectory, but the semantic encoder cannot tell two instances of these classes apart: the
distances to the target and to another member of its class coincide throughout training.
For these classes the semantic metric does not test the abstraction regime.

\paragraph{Interpretation.} In the framework of Section~\ref{sec:framework}, abstraction
requires that the target differs from prior samples along broad, high-susceptibility
features, which the memory can tilt before resolving the specific ones. Strongly
out-of-distribution targets need not offer such features. When the only broad feature is a
genre whose instances are arbitrary, the regime collapses to genre acquisition followed by
locking; when the target is a composition the prior does not represent, the memory recovers
the part the prior does represent and fails to store the rest. The abstraction regime is thus
a property of targets whose content the prior can express, in line with its contraction from
model-generated targets to photographs (Appendix~\ref{app:add_ood}).

\subsection{Memory-norm growth}\label{app:add_norm}

\begin{wrapfigure}{r}{0.4\textwidth}
  \vspace{-30pt}
  \centering
  \includegraphics[width=\linewidth]{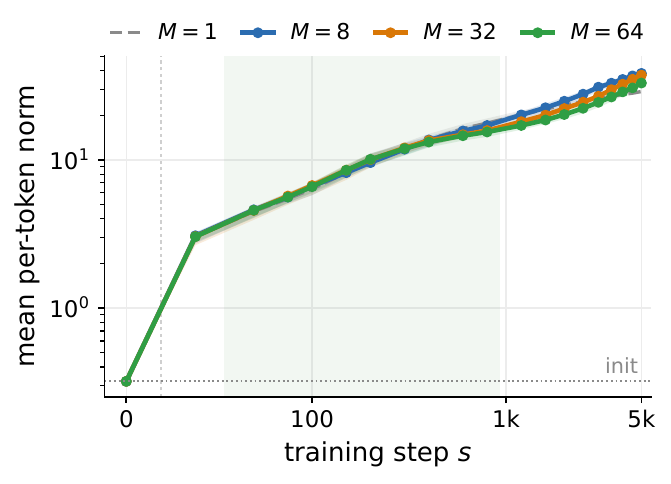}
  \vspace{-14pt}
  \caption{\textbf{Memory-norm growth.} Mean per-token norm over training (ID bank,
  $30$ runs per memory size, median with IQR band, log scale). Dotted: the initialization
  norm $\lVert\bm c\rVert\approx 0.32$.}
  \label{fig:norm_growth}
  \vspace{-25pt}
\end{wrapfigure}

Figure~\ref{fig:norm_growth} tracks the mean per-token norm of the memory vectors,
referenced from Section~\ref{sec:gist_to_verbatim}. We can notice three interesting phenomena: first of all, the norm
grows by roughly two orders of magnitude, from $0.32$ at initialization to a median of
$29$--$38$ at step $5{,}000$, without saturating; in addition to that, the trajectories are
essentially independent of memory size: $M{=}1$ through $M{=}64$ overlap throughout, so
memory size acts through the \emph{number} of tokens, not through single large tokens; finally, the
$M{=}1$ memory grows its norm exactly like the larger ones while never completing
locking (Fig.~\ref{fig:trajectory}c). 
Norm is therefore
necessary but not sufficient for locking: the two experiments below separate the
role of the memory's magnitude from that of its direction.

\subsubsection{Norm rescaling}\label{app:norm_rescale}

We take the trained memory of one in-distribution target at two checkpoints, in the
abstraction phase (step $300$, per-token norm $12$) and after locking (step
$5{,}000$, norm $37$), rescale the whole token matrix by a scalar, and sample with
the protocol of Appendix~\ref{app:implementation}
(Fig.~\ref{fig:app_norm_rescale}). Scaling the abstraction-phase memory up to the
locking norm does not produce locking: the samples lose even the semantic content
of the target. Scaling the locked memory down removes locking at once; at half its norm the samples recover the diversity and overall appearance of the abstraction phase, although the identity is less stable than in the memory trained to that stage; at the norm of the
step-$300$ memory they fall back toward the prior, whereas the memory actually
trained to that norm produces clean abstraction samples. 

\begin{figure}[h]
  \centering
  \includegraphics[width=0.85\textwidth]{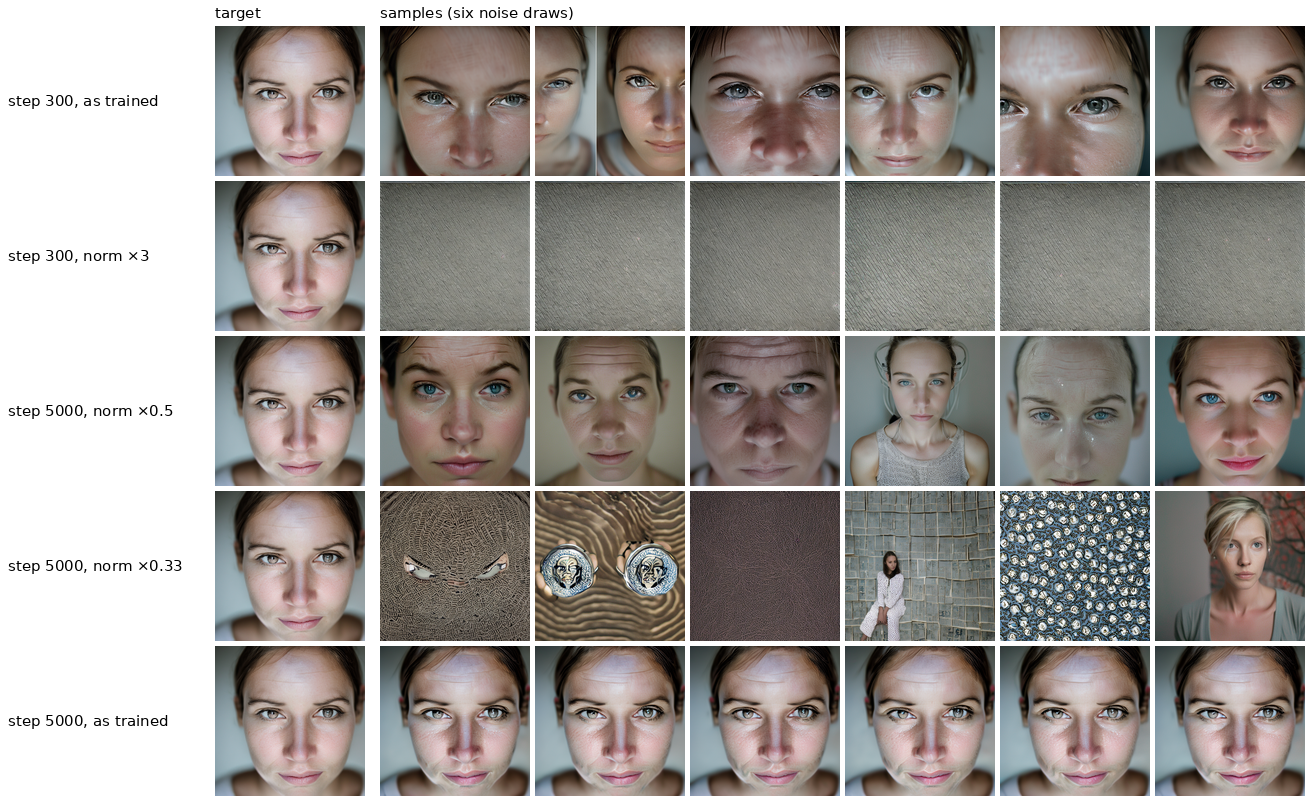}
  \caption{\textbf{Norm rescaling.} One in-distribution target, $M{=}8$, six noise
  draws per row. Rows: the step-$300$ memory as trained; the same memory scaled to
  the norm it reaches at step $5{,}000$ ($\times 3$); the step-$5{,}000$ memory at
  half its norm; the same memory scaled to the step-$300$ norm ($\times 0.33$);
  the step-$5{,}000$ memory as trained.}
  \label{fig:app_norm_rescale}
\end{figure}

\subsubsection{Fixing the norm through training}\label{app:norm_pinned}

\begin{figure}[h]
  \centering
  \includegraphics[width=\textwidth]{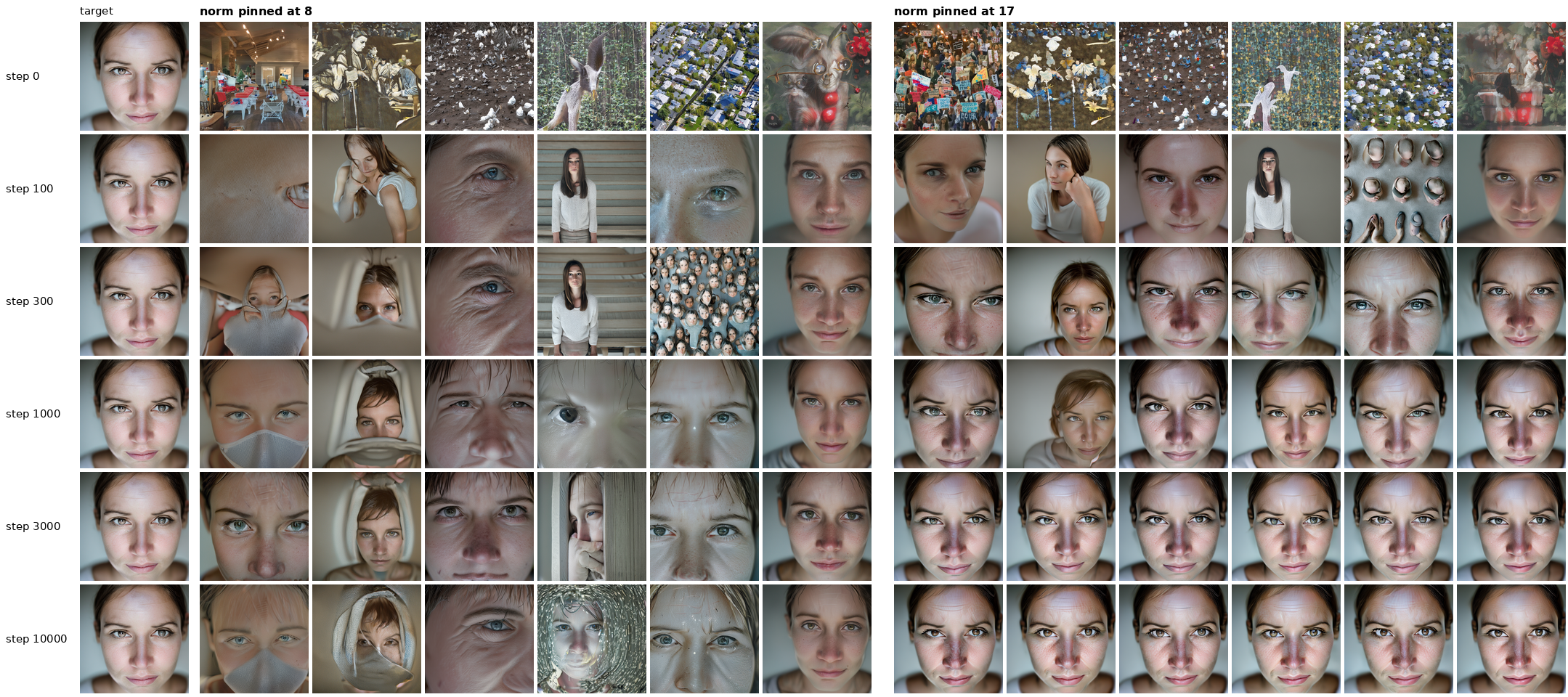}
  \caption{\textbf{Pinned norm.} The same target and $M{=}8$ as
  Fig.~\ref{fig:app_norm_rescale}, trained with the per-token norm fixed at $8$
  (left) or $17$ (right); rows are checkpoints, columns six noise draws. At norm
  $8$ the memory never locks in $10{,}000$ steps; at norm $17$ it locks by step
  $1{,}000$.}
  \label{fig:app_norm_pinned}
\end{figure}

We also train with the per-token norm pinned for the whole run, so that only the
directions of the tokens can learn (Fig.~\ref{fig:app_norm_pinned}). At norm $8$
the memory acquires the identity within $100$ steps and then stays in abstraction
for $10{,}000$ steps, never locking. At norm $17$, the norm of the text-encoder
tokens the cross-attention normally receives, it passes through the same
abstraction phase and locks by step $1{,}000$, at about half the norm free
training reaches (Fig.~\ref{fig:norm_growth}). 

\subsection{Memorizing through Low Rank Adaptation}\label{app:LoRA}
\begin{figure}[t]
    \centering
    \includegraphics[width=\linewidth]{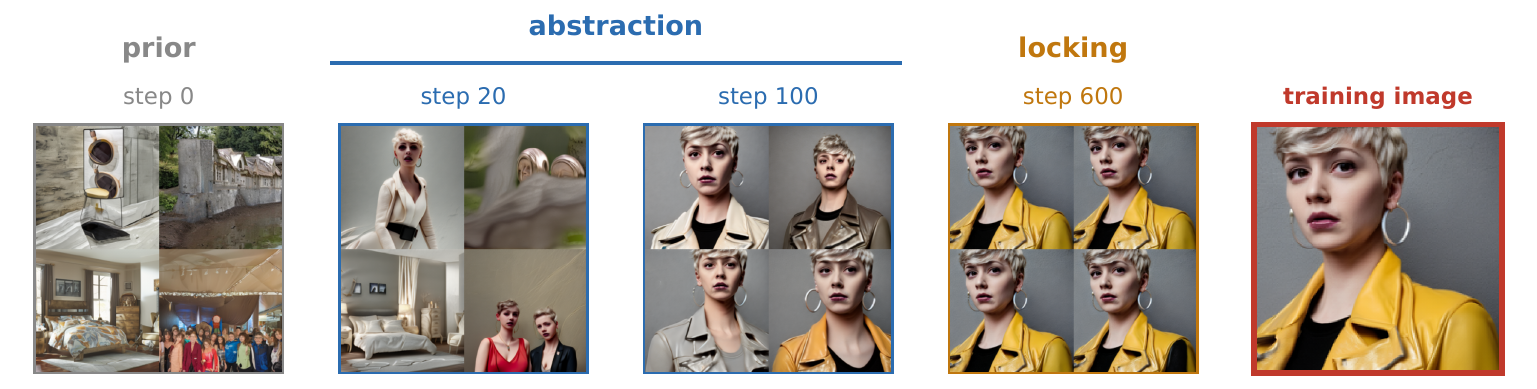}
    \vspace{-10pt}
    \caption{\textbf{Abstraction phase with LoRA}: A single SD2.1 native image, generated with a curated caption, is then memorized via a rank-1 correction of the UNet's weights (cross-attention excluded). \emph{Abstraction} then \emph{locking} emerge, in analogy to the experiments reported in the Main Text.}
    \label{fig:lora}
\end{figure}
We study an alternative route to memorization in SD2.1, closer to the implementation in \citep{ruiz2023dreambooth}. Instead of adding a parallel memory cross-attention pathway, we keep the architecture of SD2.1 unaltered, but we learn a lightweight correction to the weights of SD2.1's UNet. The optimization objective is still a single image denoising objective analogous to \eqref{eq:memory_objective}, but now the optimization variables are the LoRA parameters. 

In particular, here we consider a rank-1 correction to all weights of the UNet excluding the text-cross attention weights. This exclusion provides an ablation on the necessity of a direct interaction with the text semantic space. In Fig.~\ref{fig:lora}, we show that an abstraction phase still emerges, similar to that induced via the conditioning strategy. 

\subsection{Alternative semantics and surface metrics}\label{app:moremetrics}

\begin{figure}[h]
  \centering
  \includegraphics[width=\textwidth]{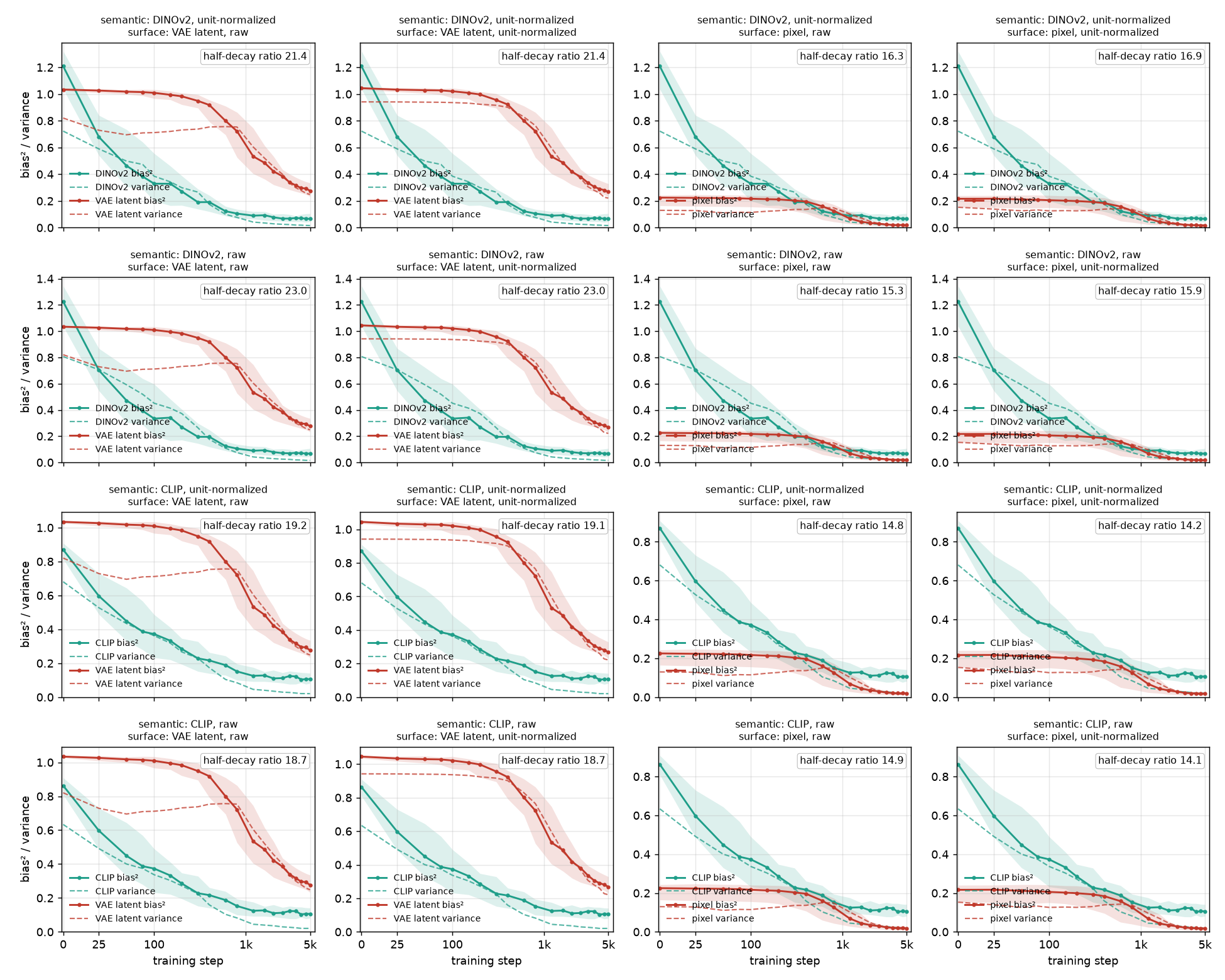}
  \caption{\textbf{All semantic/surface pairings.} Rows: semantic metric
  (DINOv2 cos, DINOv2 raw, CLIP cos, CLIP raw); columns: surface metric (VAE raw,
  VAE cos, pixel raw, pixel cos). Solid: bias$^2$; dashed: variance; median over
  the $30$ in-distribution memories at $M{=}8$, bands are interquartile ranges.
  Same protocol as Fig.~\ref{fig:trajectory}b, whose pairing is the top-left
  panel.}
  \label{fig:app_pairings}
\end{figure}

Figure~\ref{fig:trajectory}b pairs one semantic representation (unit-normalized
DINOv2) with one surface representation (raw VAE latent). Here we repeat the
analysis for two semantic embeddings (DINOv2, CLIP) and two surface
representations (VAE latent, $256^2$ pixels), each used either
\emph{unit-normalized}, with every vector rescaled to unit length before
Eq.~\eqref{eq:bias_variance}, or \emph{raw}, with the vectors as they are; in both
cases bias$^2$ and variance are divided by the squared norm of the target
representation as in Eq.~\eqref{eq:bias_variance}. Figure~\ref{fig:app_pairings}
shows the $16$ pairings on the in-distribution bank at $M{=}8$. The ordering is
the same in every panel: the semantic bias$^2$ falls early while the surface
bias$^2$ stays flat and falls late.

\subsection{Variability across training runs}\label{app:restarts}

\begin{figure}[h]
  \centering
  \includegraphics[width=\textwidth]{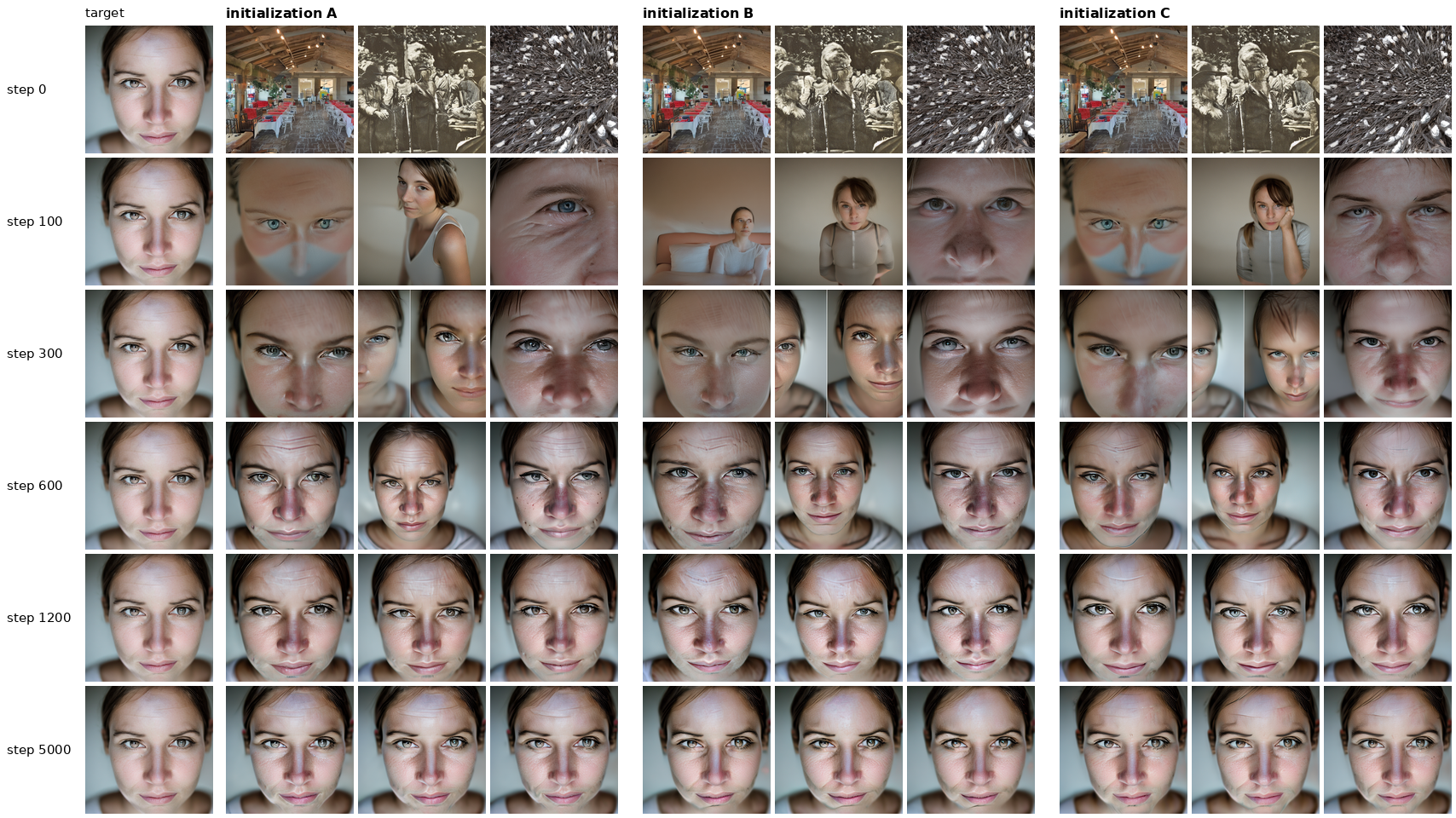}
  \caption{\textbf{Three training runs of the same memory.} One in-distribution
  target, $M{=}8$, three independent initializations (A is the run of
  Fig.~\ref{fig:trajectory}); rows are checkpoints, columns the same three noise
  draws for every run.}
  \label{fig:app_restarts}
\end{figure}

To check that the trajectory does not depend on the particular optimization run,
we retrained five face memories at $M{=}8$ from two further random
initializations (Fig.~\ref{fig:app_restarts}). The order of acquisition holds in
every run, and the ratio of surface to semantic half-decay steps varies roughly by a factor $1.3$ in median. Runs differ only in the intermediate renderings of
individual noise draws, not in when the phases occur or where the memory ends.

\subsection{Controls for the functional probes}
\label{app:controls}

This section complements the three functional probes of Section~\ref{sec:func-fidel} with controls
that test whether each effect is carried by the \emph{content} of the learned memory, rather than by
its norm, by the frozen generative prior, or by artifacts of the evaluation pipeline. Unless stated otherwise,
each probe follows the scoring protocol of
Appendices~\ref{app:exp_edit}--\ref{app:exp_retrieval}, and only the conditioning is changed.

\subsubsection{Editability}
\label{app:controls_edit}

All variants below use the memory-conditioned reversal of Appendix~\ref{app:exp_edit}: the edited image is
noised to $t_{\mathrm{switch}}$ and denoised with $\bm C^{(s)}$, using the same identities, memory
checkpoints and stochastic DDIM sampler ($\eta=1$, CFG $w=3$, 50 steps, $K=2$ samples per target--edit
pair).

\begin{figure}[t]
  \centering
  \begin{minipage}{0.325\linewidth}\centering
    \includegraphics[width=\linewidth]{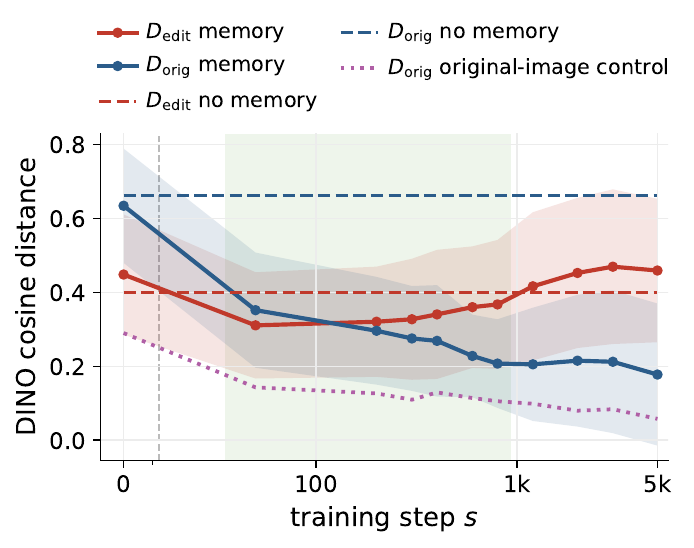}\\[-1mm]{\small (a)}
  \end{minipage}\hfill
  \begin{minipage}{0.325\linewidth}\centering
    \includegraphics[width=\linewidth]{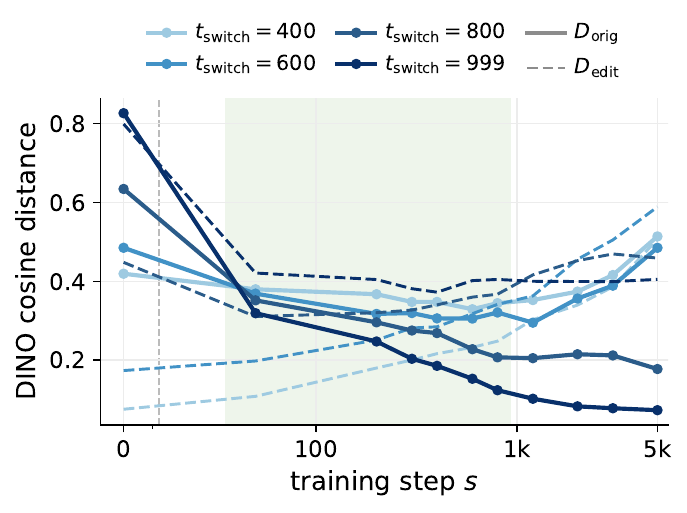}\\[-1mm]{\small (b)}
  \end{minipage}\hfill
  \begin{minipage}{0.325\linewidth}\centering
    \includegraphics[width=\linewidth]{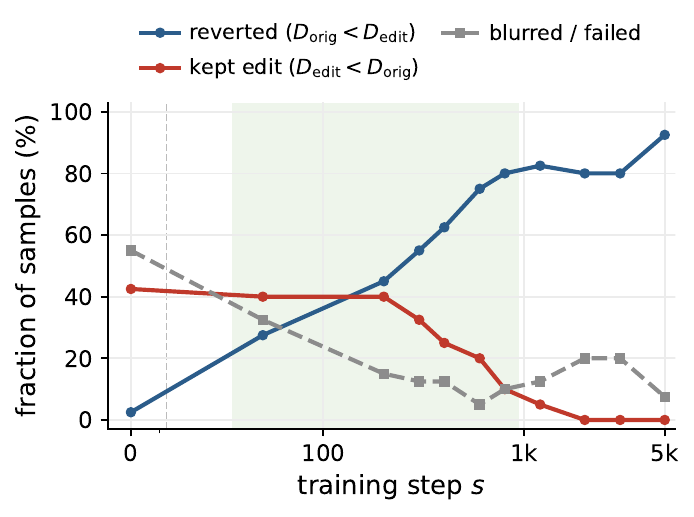}\\[-1mm]{\small (c)}
  \end{minipage}
  \caption{\textbf{Editability controls.}
  (a)~$D_{\mathrm{edit}}$ (red) and $D_{\mathrm{orig}}$ (blue) for memory-conditioned reversal at
  $t_{\mathrm{switch}}=800$ (solid; mean $\pm$ std over 20 target--edit pairs), for the no-memory control
  (dashed), and $D_{\mathrm{orig}}$ for the original-image control (dotted).
  (b)~$D_{\mathrm{orig}}$ (solid) and $D_{\mathrm{edit}}$ (dashed) for four noising levels.
  (c)~Fraction of outputs that revert to the stored target, keep the edit, or are blurred/failed
  (sharpness $<100$, or both distances $>0.5$).}
  \label{fig:controls_edit}
\end{figure}

\paragraph{No-memory control (Fig.~\ref{fig:controls_edit}a, dashed).}
We apply the identical noising and denoising but replace $\bm C^{(s)}$ with the unconditional branch,
which isolates what the frozen prior and the VAE contribute on their own. The output then stays at a fixed
distance from both images ($D_{\mathrm{edit}}=0.40$, $D_{\mathrm{orig}}=0.66$), so the prior alone does not
reintroduce the stored identity. With the memory, the output is closer to the edit than this baseline
early in training ($D_{\mathrm{edit}}=0.31$ at $s=50$), and then moves away from the edit ($0.46$ at
$s=5\mathrm{k}$) and towards the original ($D_{\mathrm{orig}}$ from $0.63$ at $s=0$ to $0.21$ at $s=800$
and $0.18$ at $s=5\mathrm{k}$). The trajectory therefore requires the learned memory.

\paragraph{Original-image control (Fig.~\ref{fig:controls_edit}a, dotted).}
We noise the unedited original $x_{\mathrm{orig}}$ instead of the edit and denoise it with $\bm C^{(s)}$
under identical settings, which measures the distortion floor of the encoding and renoising pipeline. The
output stays close to the original, with $D_{\mathrm{orig}}$ falling from $0.29$ before training to $0.06$
at $s=5\mathrm{k}$. The pipeline can thus reconstruct accurately, and a general failure of the VAE or the
noising procedure cannot explain the loss of similarity to the edit.

\paragraph{Noise level (Fig.~\ref{fig:controls_edit}b).}
We repeat the reversal at $t_{\mathrm{switch}}\in\{400,600,800,999\}$, including substantially milder
noising than the default $t_{\mathrm{switch}}=800$. At all four levels, late training moves the output
away from the edit. Even at the mildest setting ($t_{\mathrm{switch}}=400$), where the edit is initially
preserved closely ($D_{\mathrm{edit}}=0.08$ at $s=0$), $D_{\mathrm{edit}}$ rises to $0.51$ at
$s=5\mathrm{k}$ ($0.17\rightarrow0.59$ at $t_{\mathrm{switch}}=600$). The loss of editability therefore does
not depend on unusually strong noise. Reversion towards the original, in contrast, depends on the noise
level. It is clearest at strong noise ($D_{\mathrm{orig}}$ from $0.83$ to $0.07$ at
$t_{\mathrm{switch}}=999$, and from $0.63$ to $0.18$ at $800$). At milder levels, $D_{\mathrm{orig}}$ first
decreases ($0.42\rightarrow0.34$ at $t_{\mathrm{switch}}=400$, $s=800$) and then rises together with
$D_{\mathrm{edit}}$ late in training ($0.51$ at $s=5\mathrm{k}$). When little noise is added, late memories
therefore pull the output away from both images rather than cleanly back to the stored target.

\paragraph{Reversion versus failure (Fig.~\ref{fig:controls_edit}c).}
$D_{\mathrm{edit}}$ alone cannot distinguish reversion from failure: an output can be far from the edit
because it returned to the stored target, or because it is blurred, collapsed, or depicts neither image.
$D_{\mathrm{orig}}$ resolves this ambiguity. Reversion requires the output to move \emph{towards}
the stored target ($D_{\mathrm{orig}}$ decreasing, and $D_{\mathrm{orig}}<D_{\mathrm{edit}}$), whereas a
failed reconstruction stays far from both images. We additionally measure image sharpness (variance of the
Laplacian of the grayscale output), independently of both distances, and classify every output as
\emph{reverted} ($D_{\mathrm{orig}}<D_{\mathrm{edit}}$), \emph{kept edit} ($D_{\mathrm{edit}}<D_{\mathrm{orig}}$),
or \emph{blurred/failed} (sharpness $<100$, or both $D_{\mathrm{orig}}>0.5$ and $D_{\mathrm{edit}}>0.5$).
Sharpness of the memory-conditioned outputs does not drop over training (median $206$ at $s=0$, $249$ at
$s=800$, $283$ at $s=5\mathrm{k}$), and sharpness of the original-image control increases
($149\rightarrow357$). Neither shows the breakdown into blurred, unrecognizable images that a
generation-failure explanation would predict. Accordingly, reverted outputs rise from $3\%$ at $s=0$ to
$80\%$ at $s=800$ and $93\%$ at $s=5\mathrm{k}$, and kept-edit outputs vanish after $s\approx2\mathrm{k}$.
Blurred/failed outputs remain a minority ($\le20\%$ after the first $50$ steps; $7.5\%$ at
$s=5\mathrm{k}$) and are most frequent at $s=0$, where the memory carries no identity information. The loss
of edit similarity is therefore true reversion, not degradation.

\subsubsection{One-shot recognition}
\label{app:controls_recognition}

We evaluate $30$ identities with $10$ held-out pose, expression or scene edits each ($300$ queries),
classified among all $30$ identity memories (chance $1/30$). All variants share the query latents and the
timestep--noise pairs, so only the conditioning of each candidate changes. Learned memories classify $77\%$
of queries correctly early in training ($s=25$), falling to $12\%$ at $s=5\mathrm{k}$.

\begin{figure}[t]
  \centering
  \begin{minipage}{0.325\linewidth}\centering
    \includegraphics[width=\linewidth]{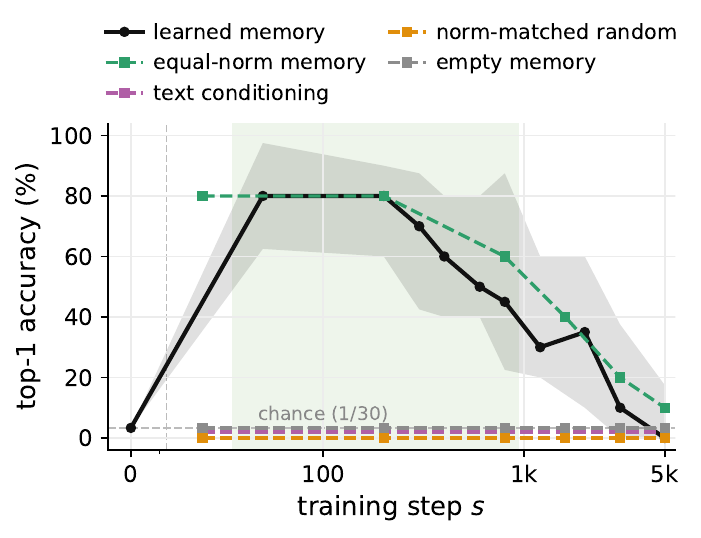}\\[-1mm]{\small (a)}
  \end{minipage}\hfill
  \begin{minipage}{0.325\linewidth}\centering
    \includegraphics[width=\linewidth]{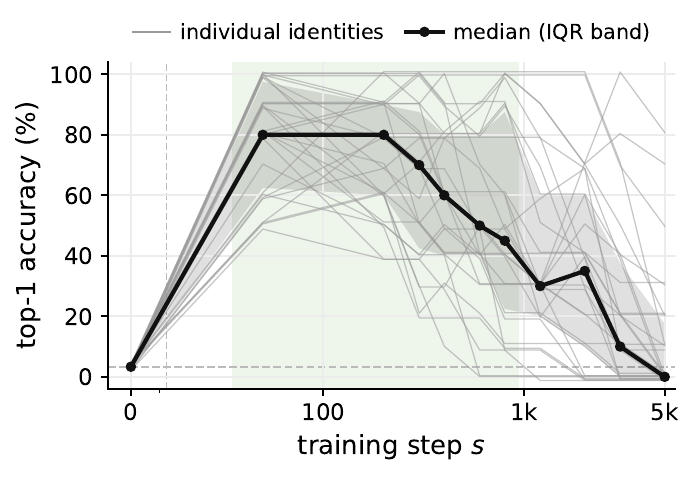}\\[-1mm]{\small (b)}
  \end{minipage}\hfill
  \begin{minipage}{0.325\linewidth}\centering
    \includegraphics[width=\linewidth]{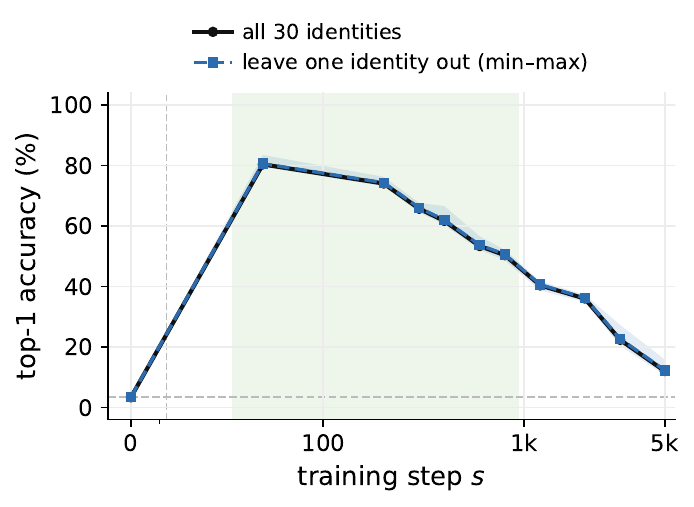}\\[-1mm]{\small (c)}
  \end{minipage}\\[2mm]
  \begin{minipage}{0.3\linewidth}\centering
    \includegraphics[width=\linewidth]{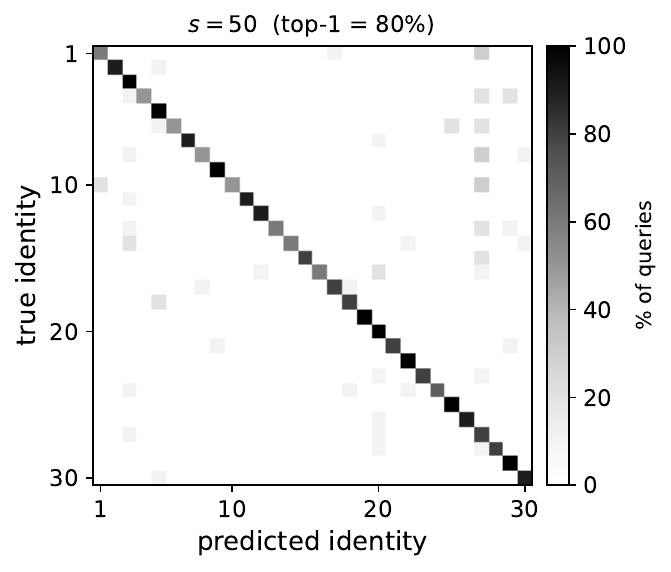}\\[-1mm]{\small (d)}
  \end{minipage}\hspace{6mm}
  \begin{minipage}{0.3\linewidth}\centering
    \includegraphics[width=\linewidth]{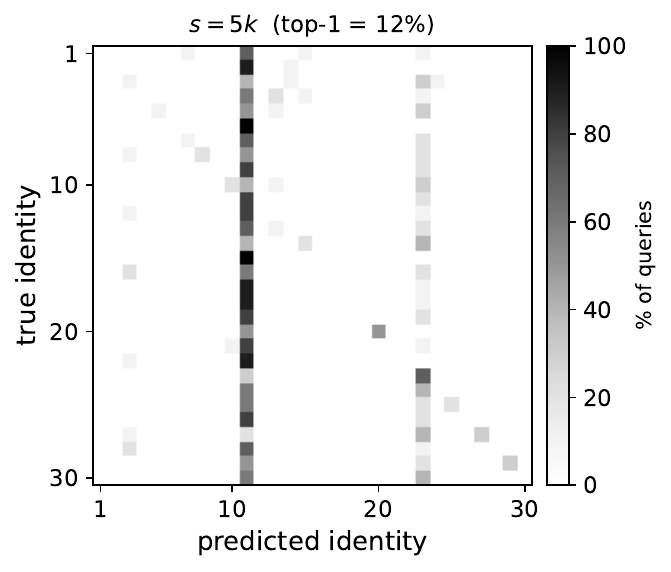}\\[-1mm]{\small (e)}
  \end{minipage}
  \caption{\textbf{One-shot recognition controls} (30 identities, 300 held-out queries).
  (a)~Median per-identity top-1 accuracy of learned memories (IQR band) versus empty, norm-matched random,
  equal-norm, and text conditioning.
  (b)~Per-identity top-1 trajectories (grey) with median and IQR.
  (c)~Leave-one-identity-out accuracy (mean and min--max over 30 folds) versus the full gallery.
  (d,e)~Confusion matrices at $s=50$ and $s=5\mathrm{k}$ (rows: true identity; entries: \% of that
  identity's queries).}
  \label{fig:controls_recognition}
\end{figure}

\paragraph{Content-free conditioning (Fig.~\ref{fig:controls_recognition}a).}
We replace the learned memories with three alternatives:
(i)~an \emph{empty} (all-zero) memory for every identity, a pipeline sanity check since all candidates
then score identically;
(ii)~a \emph{norm-matched random} memory per identity, a Gaussian vector rescaled to that identity's
learned-memory norm at the same checkpoint, i.e.\ distinguishable but content-free;
(iii)~\emph{text conditioning}, where the memory is switched off and each identity is represented by its
detailed caption (Appendix~\ref{app:exp_retrieval}) encoded by the frozen CLIP text encoder.
Empty memories yield exactly chance ($3.3\%$) and norm-matched random memories $1.3$--$4.7\%$, so
conditioning magnitude without learned content does not reproduce the signal. Text conditioning reaches a
constant $19.3\%$: captions carry some identifying information, but much less than early visual memories.

\paragraph{Equal-norm memories (Fig.~\ref{fig:controls_recognition}a).}
Across $40$ identities, memory norms vary little within a checkpoint (coefficient of variation
$6$--$11\%$) but grow substantially over training (mean $8.5$ at $s=25$ to $109$ at $s=5\mathrm{k}$). To
test whether assignments favour identities with larger norms, we rescale the learned memory directions to a
common norm (the gallery mean at each checkpoint). Both the early recognition and the decline remain
intact ($76.3\%$ vs.\ $77.3\%$ at $s=25$, $83.0\%$ vs.\ $74.0\%$ at $s=200$, $16.7\%$ vs.\ $12.0\%$ at
$s=5\mathrm{k}$), so differences in conditioning norm do not drive the result.

\paragraph{Identity-level uncertainty (Fig.~\ref{fig:controls_recognition}b).}
We report each identity's accuracy over its own $10$ queries. Recognition extends across the identity set:
at $s=200$ every identity gets between $4/10$ and $10/10$ queries correct, so the aggregate is not driven by
a few easy identities. At $s=5\mathrm{k}$ the decline is equally general: $18$ of $30$ identities are at zero
accuracy.

\paragraph{Leave-one-identity-out stability (Fig.~\ref{fig:controls_recognition}c).}
We remove each identity in turn from both gallery and queries and classify the remaining queries among
the remaining $29$ identities. Accuracy changes very little: at $s=200$ it is $74.3\pm0.7\%$ across folds
against $74.0\%$ for the full gallery, and across all checkpoints no fold deviates from the full-gallery
value by more than $4.9$ points.

\paragraph{Confusion structure (Fig.~\ref{fig:controls_recognition}d,e).}
At $s=50$ the diagonal accounts for $241/300$ queries, and the remaining $59$ errors are spread over several
identities rather than concentrated in a single confusable pair. At $s=5\mathrm{k}$ predictions collapse
onto a few identities, and a single identity attracts $199$ of $300$ queries. The decline is thus driven by
a few late memories that fit all queries well, not by uniform noise.

\begin{figure}[t]
  \centering
  \begin{minipage}{0.49\linewidth}\centering
    \includegraphics[width=\linewidth]{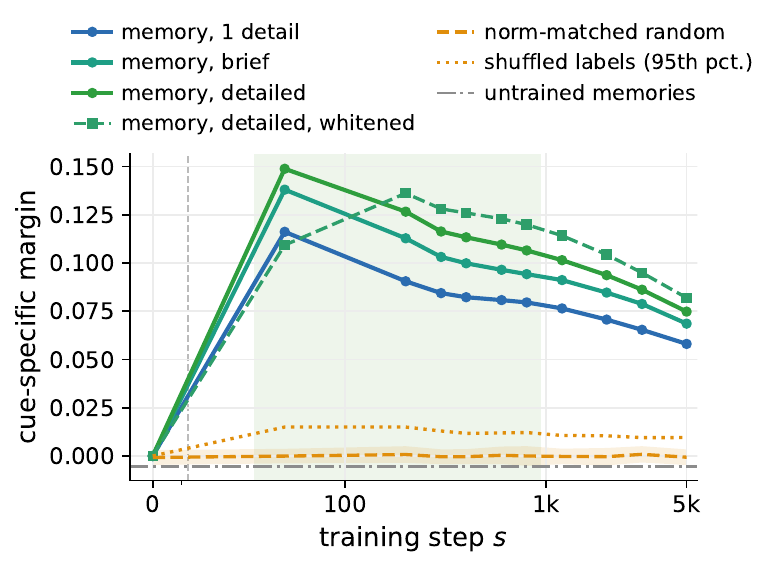}\\[-1mm]{\small (a)}
  \end{minipage}\hfill
  \begin{minipage}{0.49\linewidth}\centering
    \includegraphics[width=\linewidth]{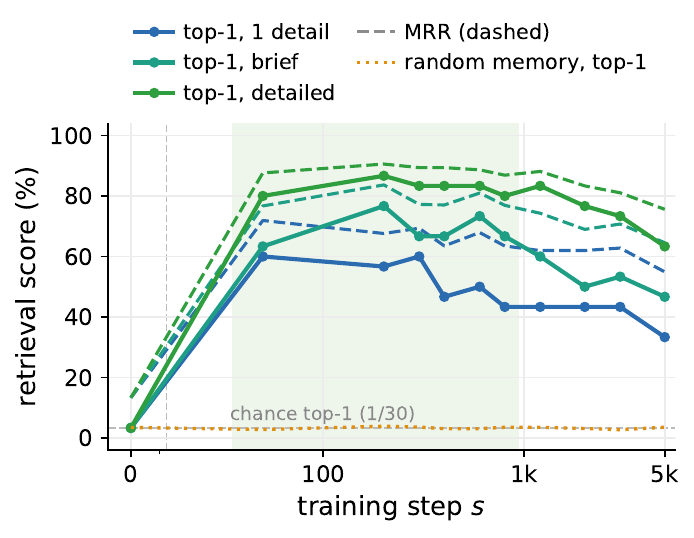}\\[-1mm]{\small (b)}
  \end{minipage}\\[3mm]
  \begin{minipage}{0.49\linewidth}\centering
    \includegraphics[width=\linewidth]{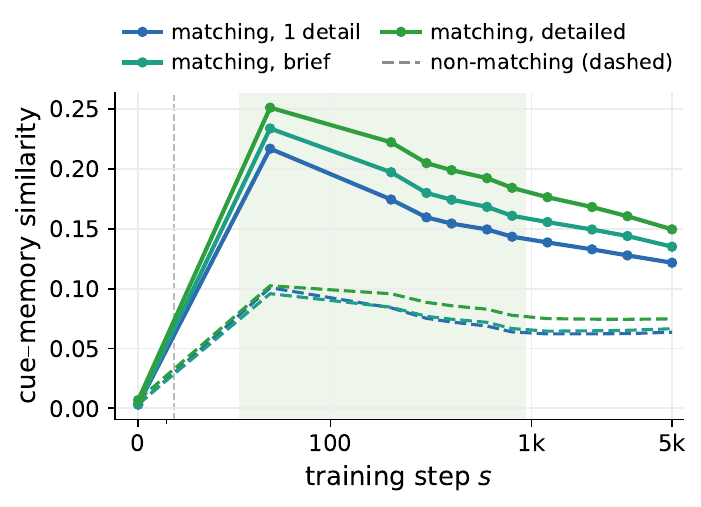}\\[-1mm]{\small (c)}
  \end{minipage}\hfill
  \begin{minipage}{0.49\linewidth}\centering
    \includegraphics[width=\linewidth]{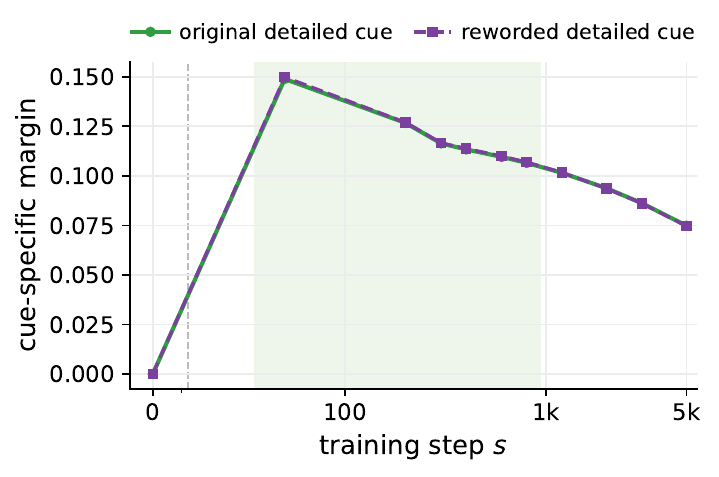}\\[-1mm]{\small (d)}
  \end{minipage}
  \caption{\textbf{Text--memory retrieval controls} (30 identities, three cue levels).
  (a)~Cue-specific margin of learned memories, after centering and standardization (dashed), versus
  norm-matched random memories (mean $\pm$ std), untrained memories, and the shuffled-label 95th percentile.
  (b)~Top-1 accuracy (solid) and MRR (dashed) per cue level, with random memories and chance for reference.
  (c)~Matching (solid) versus non-matching (dashed) cue--memory similarity.
  (d)~Margin for the original and the reworded detailed captions.}
  \label{fig:controls_retrieval}
\end{figure}

\subsubsection{Text--memory retrieval}
\label{app:controls_retrieval}

We use the cue-specific margin of Appendix~\ref{app:exp_retrieval}: the cosine similarity of a cue to its
own memory ($\mathrm{raw}_i$) minus its mean similarity to the other identities' memories
($\mathrm{base}_i$), computed for $30$ identities and three cue levels. Learned memories reach $0.149$
(detailed), $0.138$ (brief) and $0.116$ (one detail) at $s=50$. Since cosine similarity is invariant to
vector magnitude, norm differences alone cannot produce a margin. The controls below test whether
direction, anisotropy, or associations already present at initialization could.

\paragraph{Content-free memories (Fig.~\ref{fig:controls_retrieval}a).}
We replace the learned memories with \emph{norm-matched random} memories (Gaussian vectors rescaled to
each learned memory's norm) and with \emph{untrained} memories (drawn
independently per identity from the initialization distribution of the memory tokens). Random memories
give margins indistinguishable from zero at every checkpoint and cue level ($|\text{mean}|<10^{-3}$, std
$0.003$--$0.005$), and untrained memories give $-0.005$ with chance-level retrieval. The alignment therefore
develops during learning.

\paragraph{Shuffled labels (Fig.~\ref{fig:controls_retrieval}a, dotted).}
We recompute the margin under $500$ random cue--memory permutations. At every checkpoint and cue level, the
learned-memory margin exceeds the 95th percentile of this null (at most $0.017$) by at least $0.049$. The
alignment is thus specific to the matching cue rather than a generic similarity to text.

\paragraph{Centering and standardization (Fig.~\ref{fig:controls_retrieval}a, dashed).}
Cosine similarity can be inflated by structure shared across the embedding population, such as a common
direction or coordinates with much larger variance than others. We center memories and cue embeddings on
the population mean and rescale each coordinate to unit variance (statistics pooled over all memories and
checkpoints) before computing cosines. The margin remains positive at every checkpoint and cue level
(detailed: $0.075$--$0.138$) and is comparable in size to the raw margin, so shared embedding structure does
not explain it.

\paragraph{Ranking metrics (Fig.~\ref{fig:controls_retrieval}b).}
As an independent ranking measure, we report the mean reciprocal rank (MRR) of the correct identity among
all $30$ candidates, alongside top-1 accuracy. Across checkpoints, MRR is $0.76$--$0.91$ for detailed,
$0.65$--$0.86$ for brief and $0.55$--$0.78$ for one-detail cues, against $0.13$ for random memories. Top-1
accuracy with detailed cues is $80$--$87\%$ during abstraction and falls to $63\%$ at $s=5\mathrm{k}$, against
$1/30$ chance. Both rankings agree with the margin.

\paragraph{Within- versus between-identity similarity (Fig.~\ref{fig:controls_retrieval}c).}
We report the two terms of the margin separately. Similarity to the matching caption is consistently
higher than to other identities' captions, and both decrease over training: for detailed cues, matching
similarity falls from $0.251$ to $0.150$ and non-matching from $0.102$ to $0.075$
($0.234\rightarrow0.135$ vs.\ $0.096\rightarrow0.067$ for brief, $0.217\rightarrow0.122$ vs.\
$0.101\rightarrow0.064$ for one detail). Since all candidates are faces, this tests discrimination between
identities within a single semantic category.

\paragraph{Prompt wording (Fig.~\ref{fig:controls_retrieval}d).}
We rephrase each detailed caption while preserving its content (e.g.\ ``The image shows\ldots''
$\rightarrow$ ``This is a picture of\ldots'') and recompute the scores. The margin is essentially unchanged
($0.150$ vs.\ $0.149$ at $s=50$; $0.075$ vs.\ $0.075$ at $s=5\mathrm{k}$). Top-1 accuracy moves from $87\%$
to $90\%$ at $s=200$ and is unchanged at $s=5\mathrm{k}$ ($63\%$).


\subsection{Additional qualitative examples}\label{app:add_qualitative}

The qualitative panels of the main text were selected for visual clarity. Here we show some additional examples.

\begin{figure}[t]
  \centering
  \includegraphics[width=0.9\textwidth]{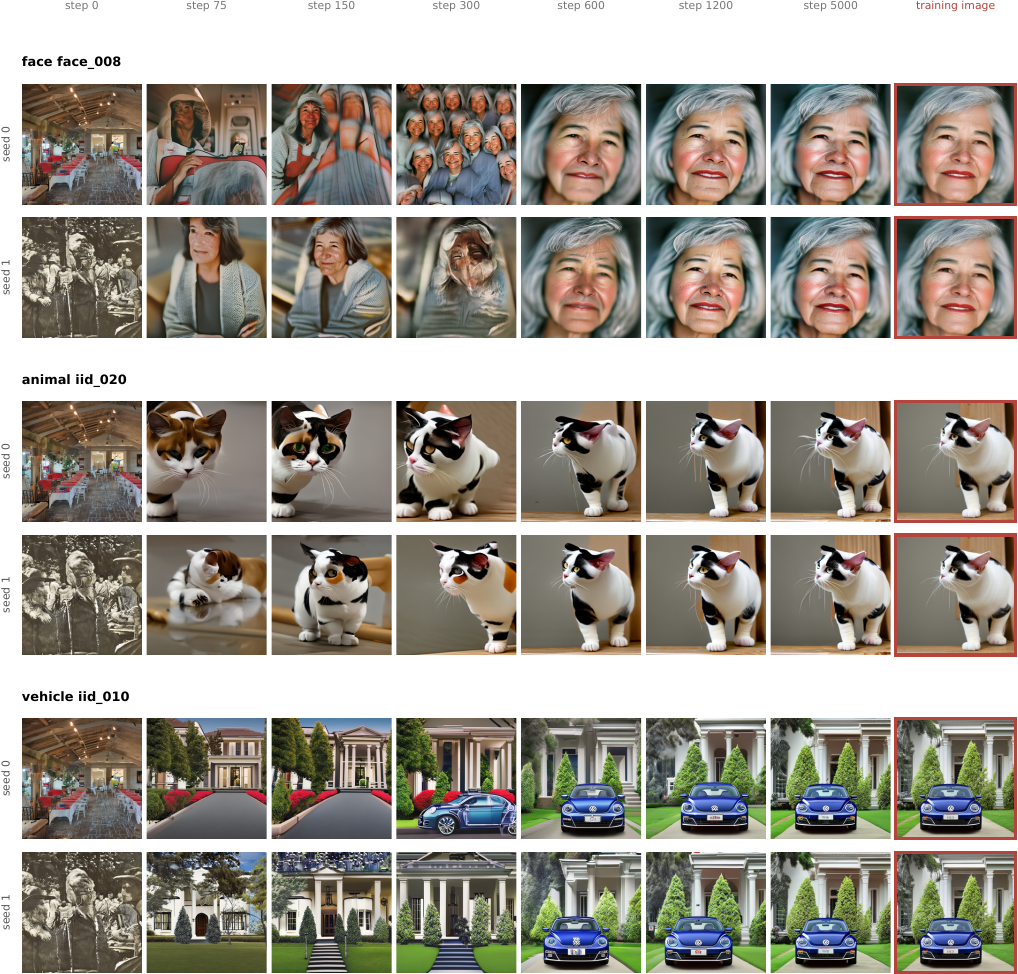}
  \caption{\textbf{Additional memorization trajectories, in-distribution.}
  Rows: the first two evaluation seeds of each run (a target's own generating seed is
  skipped). Columns: seven stored checkpoints and the training image (red frame). The
  face was drawn at random among those not shown in Fig.~\ref{fig:trajectory}. The
  animal and the vehicle are the runs whose ratio of surface to semantic half-decay steps is the median of their class.}
  \label{fig:qual_id}
\end{figure}

\paragraph{Memorization trajectories, in-distribution.} Figure~\ref{fig:qual_id} follows
three memories from the bank of Section~\ref{sec:gist_to_verbatim} across stored
checkpoints at fixed evaluation seeds. The trajectory of Fig.~\ref{fig:trajectory} is
reproduced in all three.

\paragraph{Memorization trajectories, external targets.} Figure~\ref{fig:qual_ood}
repeats the layout on the external bank of Appendix~\ref{app:add_ood}, including a plant,
a class which we have not yet shown. The same ordering holds in every block, with general details emerging before specific ones: for example, the airport is fixed before the
specific aircraft, and "roses in general" emerge before the specific target rose. 

\begin{figure}[t]
  \centering
  \includegraphics[width=\textwidth]{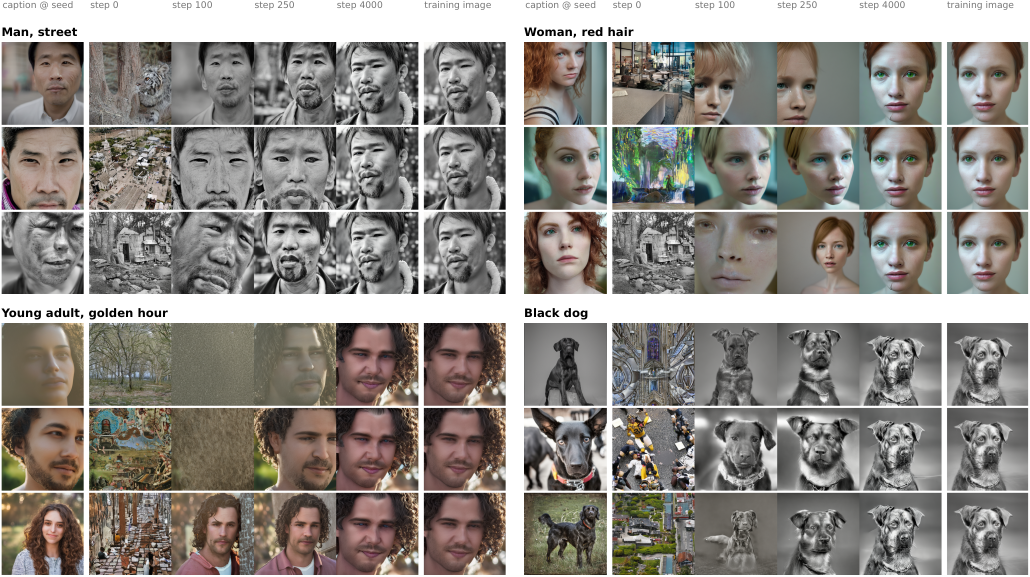}
  \caption{\textbf{Generalization at random test seeds.} Three face captions at the
  median of the twelve by peak advantage, and a black dog picture generated by SD2.1, each at three held-out seeds drawn at random.
  Columns as in Fig.~\ref{fig:generalisation}a.}
  \label{fig:qual_gen}
\end{figure}

\paragraph{Generalization at random seeds.} Figure~\ref{fig:qual_gen} shows the
generalization experiment of Section~\ref{sec:analysis} at held-out seeds drawn at random
rather than the top-ranked seeds of Fig.~\ref{fig:generalisation}a, for three face
captions and the animal caption. The results are consistent with Section~\ref{sec:analysis}: at step $100$ most seeds follow their
reference in pose and framing, some only weakly, and one has not yet left the prior at
all.

\begin{figure}[t]
    \centering
    \includegraphics[width=\linewidth]{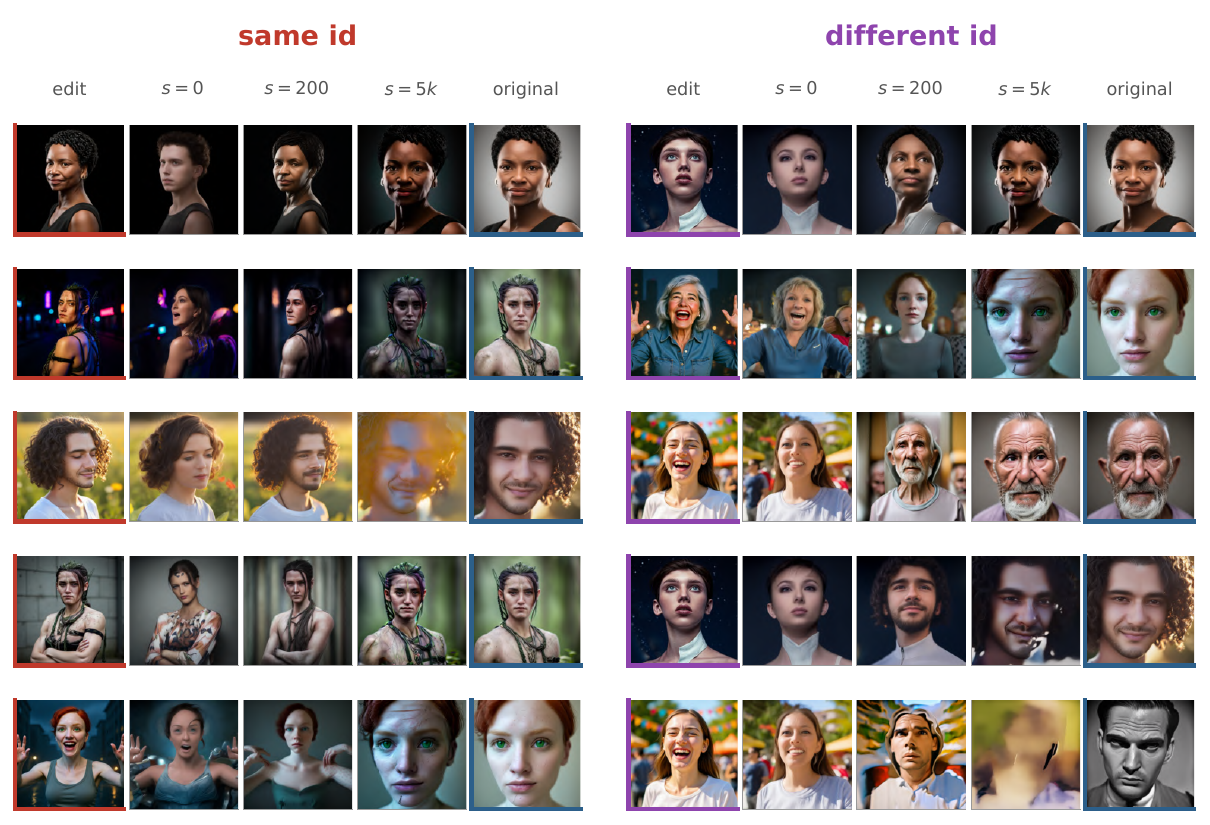}
    \caption{\textbf{Additional U-turn qualitative examples}, extending
    Fig.~\ref{fig:editability} with five more same-identity rows (left, red border) and
    five more different-identity rows (right, purple border). The protocol is the same as in
    Appendix~\ref{app:exp_edit}. Each row shows, left to right, the edited input
    $x_{\mathrm{edit}}$, the U-turn output $\hat{x}$ at the untrained memory ($s{=}0$), an
    abstract checkpoint ($s{=}200$), a memorized checkpoint ($s{=}5000$), and the stored
    target $x_{\mathrm{orig}}$ (blue border). In same-identity rows the memory belongs to
    the same person shown in the edit; in different-identity rows it belongs to a
    different person, so the output overwrites the edit with a foreign identity as
    memorization progresses. Across both groups, the edit is largely preserved at
    $s{=}0$ and $s{=}200$ and is overridden by $s{=}5000$, which is consistent with what is observed in
    Fig.~\ref{fig:editability}.}
    \label{fig:editability-extra}
\end{figure}

\paragraph{Editability, additional examples.} Figure~\ref{fig:editability-extra} extends
Fig.~\ref{fig:editability} with five more same-identity and five more different-identity
U-turn examples. The same trend holds throughout: the edit survives at $s{=}0$ and
$s{=}200$ and is increasingly overridden by $s{=}5000$. At $s{=}5000$ this override is
not always a clean reversion to the stored identity: in several rows (e.g.\ the last
one) the memory instead fails to reconstruct a coherent face at all, collapsing into a
blurred, unrecognizable output.

\begin{figure}[t]
    \centering
    \includegraphics[width=\linewidth]{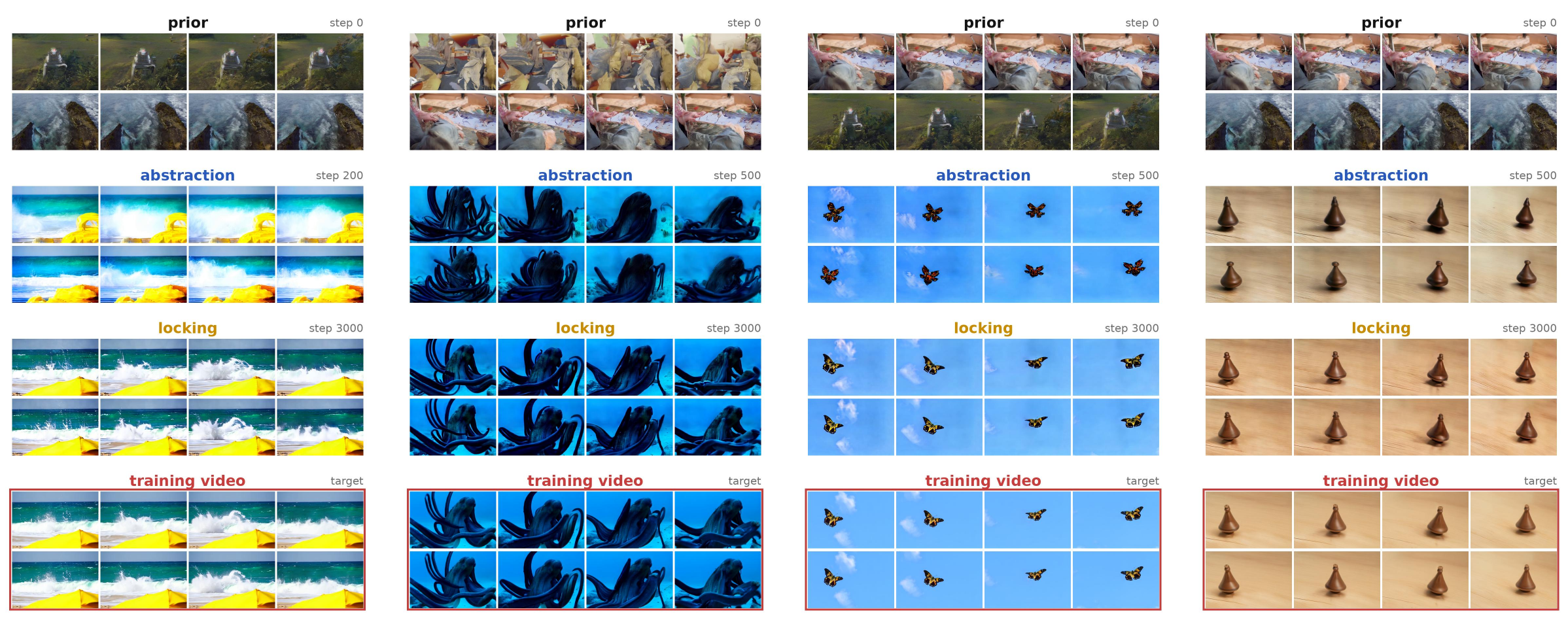}
    \caption{\textbf{Additional examples
(CogVideoX-2b).} Each column trains one memory on a single target clip on top of the
frozen video model and samples it at held-out noise seeds; columns are, left to right, breaking
\emph{waves}, an \emph{octopus} waving its tentacles, a flying \emph{butterfly}, and a
\emph{spinning top}. Rows follow the training trajectory top to bottom: draws from the prior at
initialization, the abstraction regime, locking, and the training clip (red frame). Each
block shows two seeds, each rendered as a filmstrip of four time-ordered frames. In the
abstraction regime the memory fixes the scene and its characteristic \emph{motion} (the wave
breaking, the tentacles curling, the butterfly drifting across the sky, the top spinning) while
the instance-specific appearance still varies across seeds (the umbrella's shape, the octopus'
body, the butterfly's wing pattern, the top's silhouette). Only during the locking phase does every seed
converge to the target clip.}
    \label{fig:video_examples}
\end{figure}

\paragraph{Video memories, additional examples.} Figure~\ref{fig:video_examples} extends the
CogVideoX-2b result of Fig.~\ref{fig:abstraction_modalities} with four single-clip memories
spanning distinct kinds of motion: breaking \emph{waves}, an \emph{octopus} waving its
tentacles, a flying \emph{butterfly}, and a \emph{spinning top}. The same trend holds
throughout: at initialization the memory returns unrelated draws from the prior, in the
abstraction regime it fixes the scene and its characteristic motion (the wave breaking, the
tentacles curling, the butterfly drifting, the top spinning), while the instance-specific
appearance still varies across noise seeds (the umbrella's shape, the octopus' body, the wing
pattern, the top's silhouette), and during locking every seed converges to the training clip.

\begin{figure}[t]
  \centering
  \includegraphics[width=0.72\textwidth]{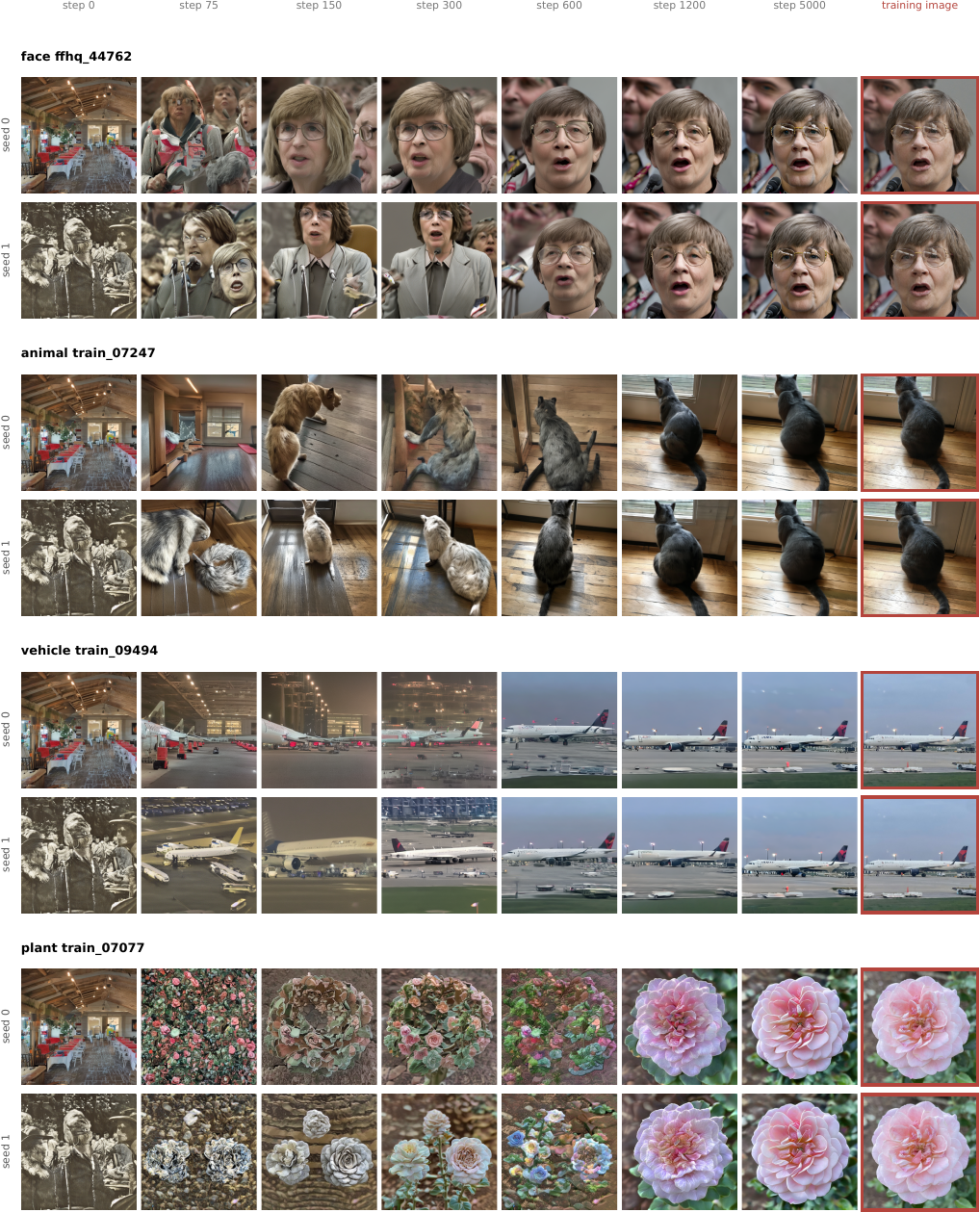}
  \caption{\textbf{Memorization trajectories, external targets.} Same layout as
  Fig.~\ref{fig:qual_id} on the bank of Appendix~\ref{app:add_ood}. The animal and the
  vehicle are the runs whose ratio of surface to semantic half-decay steps is the median of their class.}
  \label{fig:qual_ood}
\end{figure}

\end{document}